\pdfoutput=1

\documentclass[11pt]{article}

\usepackage[preprint]{acl}

\usepackage{times}
\usepackage{latexsym}
\usepackage{verbatim}

\usepackage[T1]{fontenc}

\usepackage[utf8]{inputenc}

\usepackage{microtype}

\usepackage{inconsolata}

\usepackage{graphicx}
\usepackage{tabularx}
\usepackage{setspace}
\usepackage{algpseudocode}

\usepackage[symbol]{footmisc}

\newcommand{\ul}[1]{\underline{#1}}

\newcommand\x{1.5cm}
\newcommand\y{1.7cm}

\newcommand\ti{{type\_I}}
\newcommand\tii{{type\_II}}
\newcommand\tiii{{type\_III}}

\newcommand{\sg}[1]{\textcolor{blue}{#1}}
\newcommand{\pl}[1]{\textcolor{red}{#1}}
\newcommand{\np}[1]{\textcolor{teal}{#1}}
\newcommand{\agt}[1]{\textcolor{violet}{#1}}
\newcommand{\pp}[1]{\textcolor{orange}{#1}}
\newcommand{\thm}[1]{\textcolor{blue}{#1}}
\newcommand{\loc}[1]{\textcolor{red}{#1}}

\newcommand{\smallpar}[1]{\begin{spacing}{0.7} \small #1 \end{spacing}}

\title{Are there identifiable structural parts \\ in the sentence embedding whole?}

\author{
  \textbf{Vivi Nastase\textsuperscript{1}} \and
  \textbf{Paola Merlo\textsuperscript{1,2}}
\\
  \textsuperscript{1}Idiap Research Institute, Martigny, Switzerland \\
  \textsuperscript{2}University of Geneva, Swizerland
  \\
\small{vivi.a.nastase@gmail.com, Paola.Merlo@unige.ch}
}

\begin{document}
\maketitle

\begin{abstract}
Sentence embeddings from transformer models encode in a fixed length vector much linguistic information. We explore the hypothesis that these embeddings consist of overlapping layers of information that can be separated, and on which specific types of information -- such as information about chunks and their structural and semantic properties -- can be detected. We show that this is the case using a dataset consisting of sentences with known chunk structure, and two linguistic intelligence datasets, solving which relies on detecting chunks and their grammatical number, and respectively, their semantic roles, and through analyses of the performance on the tasks and of the internal representations built during learning.
\footnote{The work was done while the authors were at the University of Geneva.}

\end{abstract}

\renewcommand*{\thefootnote}{\arabic{footnote}}

\section{Introduction}

Transformer architectures compress the information in a sentence -- morphological, grammatical, semantic, pragmatic -- into a one dimensional array of real numbers of fixed length. Sentence embeddings -- usually fine-tuned -- have proven useful for a variety of high-level language processing tasks (e.g. the GLUE tasks \citep{clark2020electra}, story continuation \cite{ippolito-etal-2020-toward}). Such higher-level tasks, however,  might not necessarily require specific structural information. Sentence embeddings built using a BiLSTM model do seem to encode a range of information, from shallow (e.g. sentence length, word order) to syntactic (e.g. tree depth, top constituent) and semantic (e.g. tense, semantic mismatches) \citep{conneau2018}. Investigation, or indeed, usage, of raw (i.e. not fine-tuned) sentence embeddings obtained from a transformer model are rare, possibly because most transformer models do not have a strong supervision signal on the sentence embedding. An investigation of the dimensions of BERT sentence embeddings using principal component analysis indicated that there is much correlation and redundancy, and that they encode more shallow information (length), rather than morphological, syntactic or semantic features \citep{nikolaev-pado-2023-universe}.
Moreover, analysis of information propagation through the model layers, and analysis of the sentence embeddings seem to show that much specialized information -- e.g. POS, syntactic structure -- while quite apparent at lower levels, gets lost towards the highest levels of the models \citep{rogers-etal-2020-primer}. 

We hypothesize that different types of information are melded together, and no longer overtly accessible in the sentence embeddings. A raw sentence embedding -- the encoding of the special [CLS]/$<s>$ token from the output of a pretrained transformer, not fine-tuned for a specific task -- consists of overlapping layers\footnote{Throughout this paper, by "layer" we mean "a stratum of information", not the layers of a transformer architecture.} of information, similarly to an audio signal that is a combination of waves of different frequencies. The various types of information from the sentence -- structural, semantic, etc. -- are encoded on some of these layers. We use a convolutional neural network to separate different layers of information in a sentence embedding, and test whether syntactic and semantic structure -- noun, verb and prepositional phrases, that may play different structural and semantic roles -- can be identified on these layers.


Understanding what kind of information the sentence embeddings encode, and how, has multiple benefits: it connects internal changes in the model parameters and structure with changes in its outputs; it contributes to verifying the robustness of models and whether or not they rely on shallow or accidental regularities in the data; it narrows down the field of search when a language model produces wrong outputs, and it helps maximize the use of training data for developing more robust models from smaller textual resources.\footnote{We will share the code and sentence data upon acceptance. The other datasets are publicly available.}

\section{Related work}

How is the information from a textual input encoded by transformers? There are two main approaches to answer this question: (i) tracing specific information from input to output through the model's various layers and components, and (ii) investigating the generated embeddings. These investigations rely on probing the models, using purposefully built data that can implement different types of testing. 

\paragraph{Tracing information through a transformer} \cite{rogers-etal-2020-primer} have shown that from the unstructured textual input, BERT \citep{devlin-etal-2019-bert} is able to infer POS, structural, entity-related, syntactic and semantic information at successively higher layers of the architecture, mirroring the classical NLP pipeline \citep{tenney-etal-2019-bert}. Further studies have shown that the information is not sharply separated, information from higher level can influence information at lower levels, such as POS in multilingual models \cite{de-vries-etal-2020-whats}, or subject-verb agreement \cite{jawahar-etal-2019-bert}. Surface syntactic and semantic information seem to be distributed throughout BERT's layers \citep{niu-etal-2022-bert,nikolaev-pado-2023-universe}. Attention is part of the process, as it helps encode various types of linguistic information \citep{rogers-etal-2020-primer,clark-etal-2019-bert}, syntactic dependencies \citep{htut2019attention}, grammatical structure \citep{luo-2021-attention}, and can contribute towards semantic role labeling \citep{tan2018deep,strubell-etal-2018-linguistically}.

\paragraph{Word embeddings} were shown to encode sentence-level information \citep{tenney-etal-2019-learn-from-context}, including syntactic structure \citep{hewitt-manning-2019-structural}, even in multilingual models \cite{chi-etal-2020-finding}. Predicate embeddings contain information about its semantic roles structure \citep{conia-navigli-2022-probing}, embeddings of nouns encode subjecthood and objecthood \citep{papadimitriou-etal-2021-deep}. The averaged token embeddings are more commonly used as {\bf sentence embeddings} (e.g. \citep{nikolaev-pado-2023-investigating}), or the special token ([CLS]/$<$s$>$) embeddings are fine-tuned for specific tasks such as story continuation \citep{ippolito-etal-2020-toward}, sentence similarity \citep{reimers-gurevych-2019-sentence}, alignment to semantic features \cite{opitz-frank-2022-sbert}. This token averaging is justifiable as the learning signal for transformer models is stronger at the token level, with a much weaker objective at the sentence level -- e.g. next sentence prediction \citep{devlin-etal-2018-bert,liu2019roberta}, sentence order prediction \citep{lan-etal-2020-albert}. Electra \citep{clark2020electra} does not either, but it relies on replaced token detection, which uses the sentence context to determine whether a (number of) token(s) in the given sentence were replaced by a generator sample. This training regime leads to sentence embeddings that perform well on the General Language Understanding Evaluation (GLUE) benchmark \citep{wang-etal-2018-glue} and Stanford Question Answering (SQuAD) dataset \citep{rajpurkar-etal-2016-squad}, or detecting verb classes \citep{yi-etal-2022-probing}.
Raw sentence embeddings also seemed to capture shallower information \cite{nikolaev-pado-2023-universe}, but \citet{nastase-merlo-2023-grammatical} show that raw sentence embeddings have internal structure that can encode grammatical sentence properties.

\paragraph{Probing models} Analysis of BERT's inner workings has been done using probing classifiers \citep{belinkov-2022-probing}, or through clustering based on the representations at the different levels \citep{jawahar-etal-2019-bert}. Probing has also been used to investigate the representations obtained from a pre-trained transformer model \citep{conneau2018}. \citet{elazar-etal-2021-amnesic} propose amnesic probing to test both whether some information is encoded, and whether it is used. VAE-based methods \citep{kingma2013vae,bowman-etal-2016-generating} have been used to detect or separate specific information from input representations.  \citet{mercatali-freitas-2021-disentangling-generative} capture discrete properties of sentences encoded with an LSTM (e.g. number and aspect of verbs) on the latent layer.  \citet{bao-etal-2019-generating} and \citet{chen-etal-2019-multi} learn to disentangle syntactic and semantic information. \citet{carvalho-etal-2023-learning} learn to disentangle the semantic roles in natural language definitions from word embeddings.

\begin{figure*}[h]
\begin{minipage}{0.45\textwidth}
\small
\begin{tabular}{llll} 
\multicolumn{4}{l}{BLM agreement problem} \\
\hline
\multicolumn{4}{c}{\sc Context Template}\\
\hline
\sg{NP-sg}& \sg{PP1-sg}& & \sg{VP-sg}  \\
\pl{NP-pl} & \sg{PP1-sg}& & \pl{VP-pl}  \\
\sg{NP-sg}& \pl{PP1-pl} & & \sg{VP-sg}  \\
\pl{NP-pl} & \pl{PP1-pl} & & \pl{VP-pl}  \\
\sg{NP-sg}& \sg{PP1-sg}& \sg{PP2-sg} &\sg{VP-sg}  \\
\pl{NP-pl} & \sg{PP1-sg}  & \sg{PP2-sg} &            \pl{VP-pl}  \\
\sg{NP-sg}   & \pl{PP1-pl} & \sg{PP2-sg} &             \sg{VP-sg}  \\
\end{tabular}

  \begin{tabular}{ll} \hline
 \multicolumn{2}{c}{{\sc Answer set} } \\ \hline
NP-sg PP1-sg et NP2 VP-sg & Coord  \\ 
{\bf NP-pl PP1-pl NP2-sg VP-pl } & correct \\
NP-sg PP1-sg VP-sg & WNA\\
NP-pl PP1-pl NP2-pl VP-sg & AE\_V \\
NP-pl PP1-sg NP2-pl VP-sg & AE\_N1 \\
NP-pl PP1-pl NP2-sg VP-sg & AE\_N2 \\
NP-pl PP1-sg PP1-sg VP-pl & WN1 \\
NP-pl PP1-pl PP2-pl VP-pl & WN2 \\
 \hline
 \end{tabular}
\end{minipage}
\hfill
\begin{minipage}{0.55\textwidth}
\small
\setlength{\tabcolsep}{3pt} 
\begin{tabular}{llll} 
\multicolumn{4}{l}{BLM verb alternation problem} \\
\hline
\multicolumn{4}{c}{\sc Context Template}\\
\hline
\np{NP-\agt{Agent}} & Verb & \np{NP-}\loc{Loc} & \pp{PP-}\thm{Theme} \\
\np{NP-\thm{Theme}} & VerbPass & \pp{PP-}\agt{Agent} \\
\np{NP-\thm{Theme}} & VerbPass & \pp{PP-}\loc{Loc} & \pp{PP-}\agt{Agent}\\
\np{NP-\thm{Theme}} & VerbPass & \pp{PP-}\loc{Loc}  \\
\np{NP-\loc{Loc}} & VerbPass & \pp{PP-}\agt{Agent} \\
\np{NP-\loc{Loc}} & VerbPass & \pp{PP-}\thm{Theme} & \pp{PP-}\agt{Agent} \\
\np{NP-\loc{Loc}} & VerbPass & \pp{PP-}\thm{Theme} \\
\end{tabular}
\begin{tabular}{ll} \hline
\multicolumn{2}{c}{\sc Answer set}\\ \hline
\bf{NP-Agent Verb NP-Theme PP-Loc} & \textsc{Correct} \\ 
NP-Agent *VerbPass NP-Theme PP-Loc & \textsc{AgentAct}\\
NP-Agent Verb NP-Theme *NP-Loc & \textsc{Alt1}\\
NP-Agent Verb *PP-Theme PP-Loc & \textsc{Alt2}\\
NP-Agent Verb *[NP-Theme PP-Loc] & \textsc{NoEmb}\\
NP-Agent Verb NP-Theme *PP-Loc & \textsc{LexPrep}\\
*NP-Theme Verb NP-Agent PP-Loc & \textsc{SSM1}\\
*NP-Loc Verb NP-Agent PP-Theme &\textsc{SSM2}\\
*NP-Theme Verb NP-Loc PP-Agent & \textsc{AASSM}\\
\hline
 \end{tabular}
\label{ALT-ATL}
\end{minipage}
\caption{Structure of two BLM problems, in terms of chunks in sentences and sequence structure.}
\label{fig:BLM_structure}
\end{figure*}

\paragraph{Data} For probing transfomers embeddings and behaviour, most approaches use datasets built by selecting, or constructing, sentences that exhibit specific structure and properties: definition sentences with annotated roles \citep{carvalho-etal-2023-learning}, sentences built according to a given template \citep{nikolaev-pado-2023-representation}, sentences with specific structures for investigating different tasks, in particular SentEval \cite{conneau-kiela-2018-senteval} \citep{jawahar-etal-2019-bert}, example sentences from FrameNet \citep{conia-navigli-2022-probing}, a dataset with multi-level structure inspired by the Raven Progressive Matrices visual intelligence tests \citep{an-etal-2023-blm}.



\section{Data}
\label{sec:data}



Our main object of investigation are chunks, sequence of adjacent words that segment a sentence (as defined initially in \cite{abney-1992-prosodic}, \cite{collins-1997-three} and then \cite{tjong-kim-sang-buchholz-2000-introduction}. To investigate whether chunks and their properties are identifiable in sentence embeddings, we use two types of data: (i) sentences with known chunk pattern, described in Section \ref{sec:sent_descr}; (ii) two datasets with multi-level structure built for linguistic intelligence tests for language models \cite{merlo2023}, described in Section \ref{sec:BLMs}.

\subsection{Sentences}
\label{sec:sent_descr}

Sentences are built from a seed file containing noun, verb and prepositional phrases, including singular/plural variations. From these chunks, we built sentences with all (grammatically correct) combinations of \texttt{np (pp$_1$ (pp$_2$)) vp}\footnote{We use BNF notation: pp$_1$ and pp$_2$ may be included or not, pp$_2$ may be included only if pp1 is included}. For each chunk pattern $p$ of the 14 possibilities (for instance,  $p$ = "np-s pp1-s vp-s"), all corresponding sentences are collected into a set $S_p$. 

We generate an instance for each sentence $s$ from the sets $S_p$  as a triple $(in, out^+, out^-)$, where $in=s$ is the input, $out^+$ is the correct output, which is a sentence different from $s$ but having the same chunk pattern. $out^-$ are $N_{negs}$ incorrect outputs, randomly chosen from the sentences that have a chunk pattern different from $s$. The algorithm for building the data and a sample line and generated sentences are shown in appendix \ref{sec:sentencedata}.

From the generated instances, we sample uniformly, based on the pattern of the input sentence, approximately 4000 instances,  randomly split 80:20 into train:test. The train part is further split 80:20 into train:dev, resulting in a 2576:630:798 split for train:dev:test. We use a French and an English seed file and generate French and English variations of the dataset, with the same statistics.

\subsection{Blackbird Language Matrices}
\label{sec:BLMs}

Blackbird Language Matrices (BLMs) \citep{merlo2023} are language versions of the visual Raven Progressive Matrices. They are multiple-choice problems, where the input is a sequence of sentences built using specific rules, and the answer set consists of a correct answer that continues the input sequence, and several incorrect options that are built by corrupting some of the underlying generating rules of the sentences in the input sequence. In a  BLM matrix, all sentences  share a targeted linguistic phenomenon, but differ in other aspects relevant for the phenomenon in question.   Thus, BLMs, like their visual counterpart RPMs, require identifying the entities (the chunks), their relevant  attributes (their morphological or semantic properties) and their connecting operators, to find the underlying rules that guide to the correct answer.  

We use two BLM datasets, which encode two different linguistic phenomena, each in a different language: (i) BLM-AgrF -- subject verb agreement in French \citep{an-etal-2023-blm}, and (ii) BLM-s/lE -- verb alternations in English \citep{samo-etal-2023}. The structure of these datasets -- in terms of the sentence chunks and sequence structure -- is shown in Figure \ref{fig:BLM_structure}, and concrete examples are shown in appendices \ref{app:agr}, \ref{app:verbalt}.


BLM datasets also have a lexical variation dimension. There are three variants: type I -- minimal lexical variation for sentences within an instance, type II -- one word difference across the sentences within an instance, type III -- maximal lexical variation within an instance. This allows for investigations in the impact of lexical variation on learning the relevant structures to solve the problems.

We use the BLM-s/lE dataset as is. We built a variation of the BLM-AgrF \citep{an-etal-2023-blm} that separates clearly sequence-based errors (WN1 and WN2 in the agreement scheme presented in Figure \ref{fig:BLM_structure}) from other types of errors. We include erroneous answers that have correct agreement, but do not respect the pattern of the sequence, to be able to contrast linguistic errors from errors in identifying sentence parts.

\begin{table}
  \vspace{-5mm}
  \small
    \begin{tabular}{l|r|rr}
     & Subj.-verb agr. & \multicolumn{2}{c}{Verb alternations} \\ 
     &                 & ALT-ATL & ATL-ALT \\ \hline
    Type I  & 2000:252  &  2000:375 & 2000:375 \\
    Type II & 2000:4866 &  2000:1500 & 2000:1500 \\
    Type III & 2000:4869 &  2000:1500 & 2000:1500 \\ \hline
    \end{tabular}
    \caption{Train:Test statistics for the two BLM problems. }
    \label{tab:data}
\end{table}

\paragraph{Datasets statistics}
Table \ref{tab:data} shows the datasets statistics for the BLM problems. After splitting each subset 90:10 into train:test subsets, we randomly sample 2000 instances as train data. 20\% of the train data is used for development. Types I, II, III correspond to different amounts of lexical variation within a problem instance.

\section{Experiments}

We aim to determine whether specific kinds of sentence parts -- chunks -- are identifiable in transformer-based sentence embeddings. 
We approach this problem from two angles. First, using sentences and a VAE-based system, we test whether we can compress sentences into a smaller representation on the latent layer that captures information about the chunk structure of the sentence (Section \ref{sec:sentences} below). Second, to see if the chunks thus identified are being used in a separate task, we combine the compression of the sentence representation with the BLM problems, where  a crucial part of the solution lies in  identifying the structures of sentences and their sequence in the input (Section \ref{sec:parts-for-BLMs} below). 

As sentence representations, we use the embeddings of the \textit{$<s>$} character read from the last layer of the Electra \cite{clark2020electra} pretrained model\footnote{Electra pretrained model: {\it google/electra-base-discriminator}}.

\subsection{Parts in sentences} 
\label{sec:sentences}

We test whether sentence embeddings contain information about the chunk structure of the corresponding sentences by compressing them into a lower dimensional representation in a VAE-like system. 

\subsubsection{Experimental set-up}
The architecture of the sentence-level VAE is similar to a previously proposed system \cite{nastase-merlo-2023-grammatical}. 
 The encoder consists of a CNN layer with a 15x15 kernel, which is applied to a 32x24-shaped sentence embedding,\footnote{\citet{nastase-merlo-2023-grammatical} show that task-relevant information is more easily accessible in transformer-based sentence embeddings reshaped as two-dimensional arrays, which indicates that patterns are encoded periodically, and are best detected with a 15x15 kernel.}  followed by a linear layer that compresses the output of the CNN into a latent layer of size 5. The decoder is a mirror-image of the encoder, and unpacks a sampled latent vector into a 32x24 sentence representation.

An instance consists of a triple $(in, out^+, out^-)$, where $in$ is an input sentence with embedding $e_i$ and chunk structure $p$, $out^+$ is a sentence with embedding $e_j$ with  same chunk structure $p$, and $out^-$ is a set of $N_{negs}$ sentences with embeddings $e_k$, each of which has a chunk pattern different from $p$ (and different from each other). The input $e_i$ is encoded into a latent representation $z_i$, from which we sample a vector $\tilde{z}_i$, which is decoded into the output $\hat{e}_i$. We enforce that the latent encodes the structure of the input sentence by using a max-margin loss function. This loss function assigns a higher score to $e_j$ than to $e_k$, relative to $\hat{e}_i$. Recall that $e_j$ has the same chunk structure as the input $\hat{e}_i$.  

\vspace{2mm}

\noindent$loss(e_i) = \\ \mbox{~~~~~} maxmargin(\hat{e}_i,e_j, e_k) + KL(z_i||\mathcal{N}(0,1))$
\vspace{3mm}

\noindent $maxmargin(\hat{e}_i,e_j, e_k) = \\ \mbox{~~~} max(0, 1-score(\hat{e}_i, e_j) + \frac{\sum_{k=1}^{N_{negs}} score(\hat{e}_i, e_k)}{N_{negs}})$

\vspace{2mm}

The \textit{score} between two embeddings is the cosine similarity. At prediction time, the sentence from the $\{out^+\} \cup out^-$ options that has the highest score relative to the input sentence is taken as the correct answer.

\subsubsection{Analysis}

To assess whether the correct patterns of chunks are detected in sentences, we analyze the results for the experiments described in the previous section in two ways: (i) analyze the output of the system, in terms of average F1 score over three runs and confusion matrices; (ii) analyze the latent layer, to determine whether chunk patterns are encoded in the latent vectors (for instance,  latent vectors cluster according to the pattern of their corresponding sentences).

\begin{figure}[h]
        \centering
        \includegraphics[width=0.48\textwidth,trim={0 0 0 2.1cm},clip]{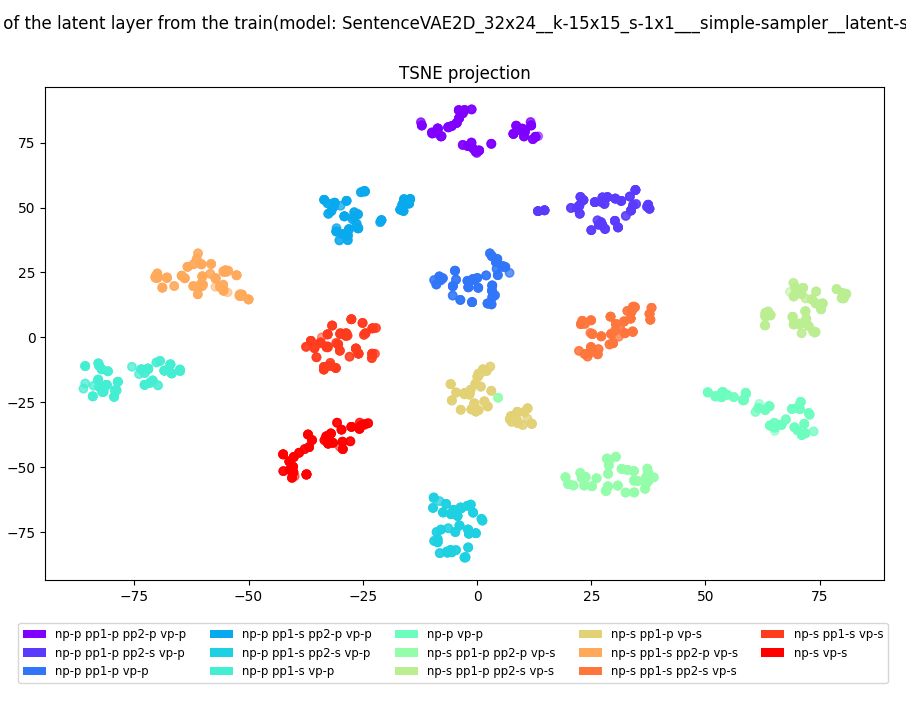}        
        
        \includegraphics[width=0.48\textwidth]{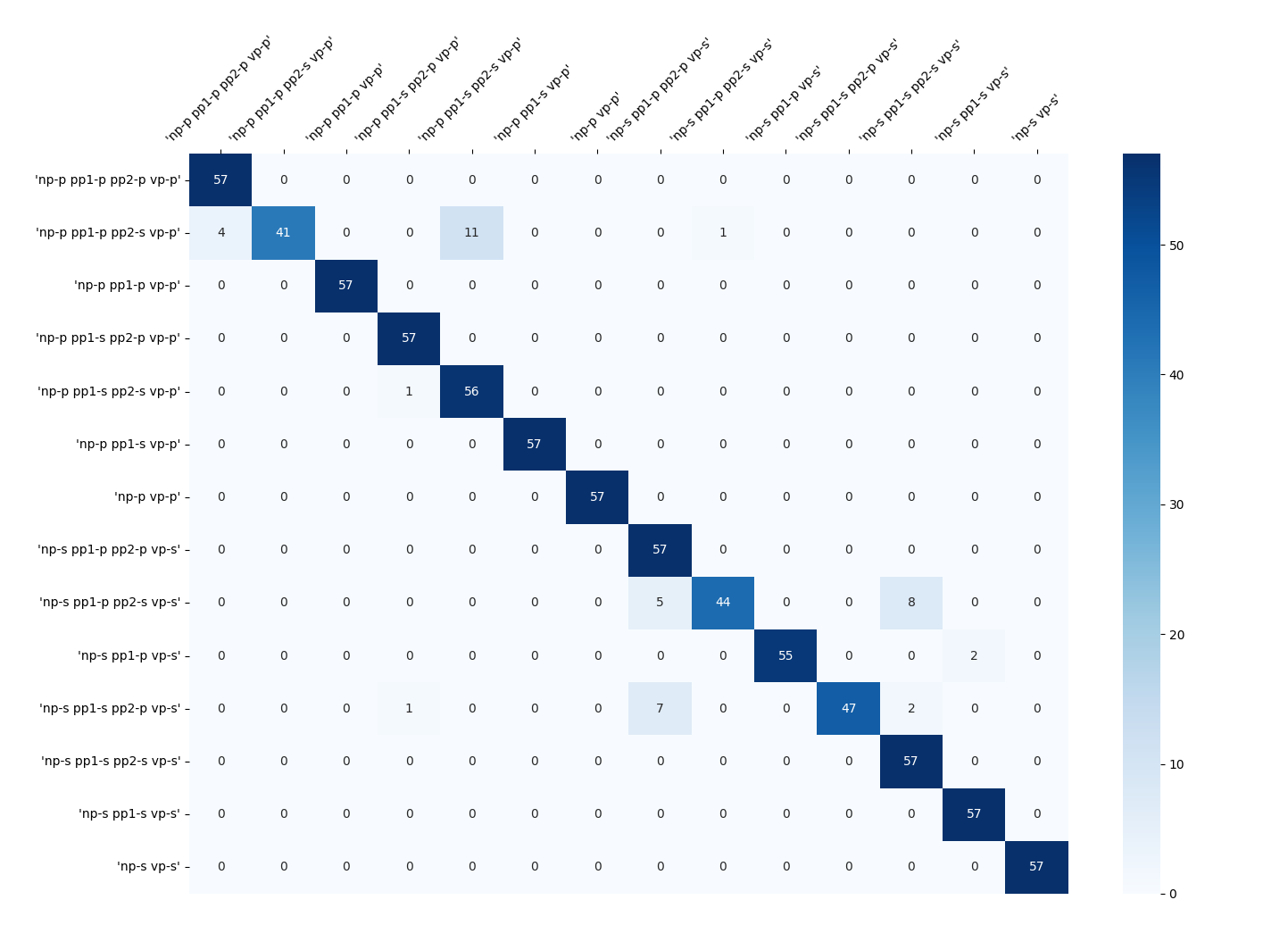}
    \caption{Chunk identification results: tSNE projections of the latent vectors for the French dataset, and confusion matrix of the system output. The results for English are similar.}
    \label{fig:partsinsentences}
\end{figure}

If we consider the multiple choice task as a binary task (Has the system built a sentence representation that is closest to the correct answer?), the system achieves an average positive class F1 score (and standard deviation) over three runs of 0.9992 (0.01) for the French dataset, and 0.997 (0.0035) for the English dataset. For added insight, for one trained model for each of the French and English data, we compute a confusion matrix, based on the pattern information for $out^+, out^-$. The results for French are presented in Figure \ref{fig:partsinsentences}. 

To check whether chunk information is present in the latent layer, we plot the projection in two dimensions of the latent vectors. The plot shows a very crisp clustering of latents that correspond to input sentences with the same chunk pattern, despite the fact that some patterns differ by only one attribute (the grammatical number) of one chunk.

\begin{figure}[h]
    \centering
    \includegraphics[width=0.49\textwidth]{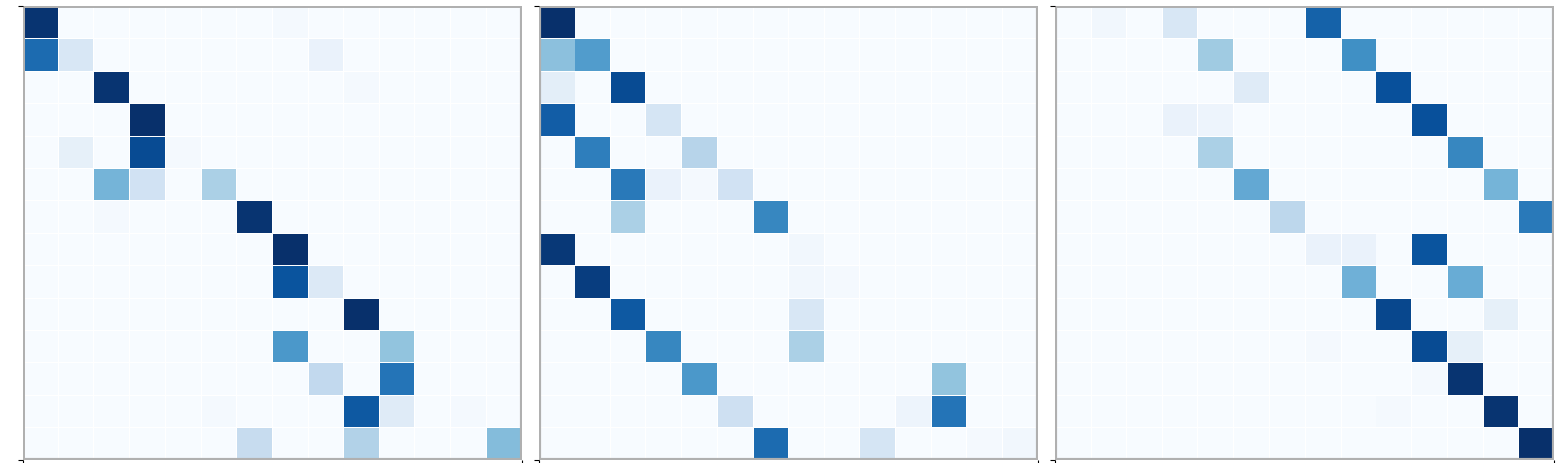}
    \captionof{figure}{The impact on reconstructing sentences with the same pattern when modifying the latent layer with values in their respective min-max range (based on the training data) -- sample confusion matrices.}
    \label{fig:sentence_latents}
\end{figure}

To understand how chunk information is encoded on the latent layer we perform latent traversals: for each instance in the test data, we modify the value of each unit in the latent layer with ten values in the min-max range of that unit, based on the training data. A sample of confusion matrices with interventions on the latent layer is shown in Figure \ref{fig:sentence_latents}. 

The confusion matrices presented as heatmaps in Figure \ref{fig:sentence_latents} (and a larger version with labels in Figure \ref{fig:confmats} in Appendix \ref{sec:confmats}) show that specific changes to the latent vectors decrease the differentiation among patterns, as expected if chunk pattern information were encoded in the latent vectors. Changes to latent 1 cause patterns that differ in the grammatical number of $pp2$ not to be distinguishable (left matrix). Changes to latent units 2 and 3 lead to the matrices 2 and 3 in the figure, where patterns that have different subject-verb grammatical number to become indistinguishable.

\subsubsection{Electra vs. BERT and RoBERTa, and the price of fine-tuning}

We have chosen Electra for the investigations presented here, because with its use of the full context of the sentence to determine whether tokens in the input have been replaced by a generator model, it provides a stronger supervision signal than BERT \citep{devlin-etal-2019-bert} and RoBERTa \citep{liu2019roberta}. To check whether the sentence embeddings produced by the three systems differ in their encoding of chunks, we ran the same experiment on the data encoded with BERT\footnote{\url{https://huggingface.co/google-bert/bert-base-multilingual-cased}} and RoBERTa\footnote{\url{https://huggingface.co/FacebookAI/xlm-roberta-base}}. In terms of F1 score on the task of reconstructing a sentence with the same chunk structure, BERT has a mean over 3 runs of 0.91 (std=0.0346), while RoBERTA has 0.8926 (std=0.0166). 


Sentence embeddings are often fine-tuned for specific tasks. We tested two sentence transformer models LaBSE and MPNet\footnote{\url{https://huggingface.co/sentence-transformers/LaBSE, https://huggingface.co/sentence-transformers/all-mpnet-base-v2}}, and obtained an F1 mean of 0.43 (std=0.0336) and 0.669 (std=0.0407) respectively. We chose LaBSE and MPNet because they are two differently tuned models --LaBSE is trained with bilingual sentence pairs with high results on a cross-language sentence retrieval task, MPNet is optimized for sentence similarity -- and their representations have the same dimensionality (768) as the transformer models we used. The low results on detecting chunk structure in sentence embeddings after this tuning indicates that in the quest of optimizing the representation of the meaning of a sentence, structural information is lost.

\subsection{Parts in sentences for BLM tasks}
\label{sec:parts-for-BLMs}

The first experiment shows that compressing sentence representations results in latent vectors containing chunk information. To test if these latent representations also contain information about chunk properties relevant to a task, we solve the BLM task. 

\subsubsection{Experimental set-up}

\begin{figure}[h]
    \centering
    \includegraphics[width=0.45\textwidth]{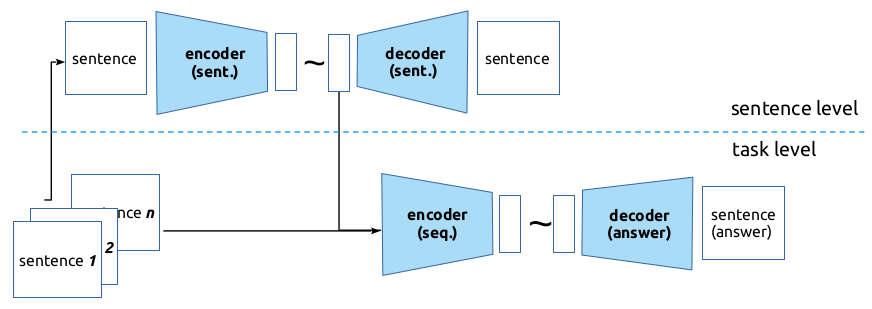}
    \caption{A two-level VAE: the sentence level learns to compress a sentence into a representation useful to solve the BLM problem on the task level.}
    \label{fig:2levelVAE}
\end{figure}

To explore how chunk information in the sentence embeddings is used in a task, we solve the BLM problems. The BLM problems encode a linguistic phenomenon in a sequence of sentences that have regular and relevant structure, which serves to emphasize and reinforce the encoded phenomenon. BLMs are inspired by Raven Progressive Matrices, whose solution has been shown to require solving two main subtasks: identifying the objects and object attributes that occur in the visual frames, and decomposing the main problem into subproblems, based on object and attribute identification, in a way that allows detecting the global pattern or underlying rules. It has also been shown that being able to solve RPMs requires being able to handle item novelty \cite{carpenter1990one}. 
We model these ingredients of the solution of a RPM/BLM explicitly by using the two-level intertwined architecture illustrated in Figure \ref{fig:2levelVAE} -- one level for detecting sentence structure, one for detecting the correct answer based on the sequence of structures and the targeted grammatical phenomenon. Item novelty is modeled through the three levels of lexicalisation (section \ref{sec:data}).

The sentence level is essentially the system described above. The representation on the latent layer is used to represent each of the sentences in the input sequence, and to solve the problem at the task level. The two layers are trained together.

An instance for a BLM problem consists of an ordered sequence $S$ of sentences, $S = \{s_i| i = 1,7\}$ as input, and an answer set $A$ with one correct answer $a_c$, and several incorrect answers $a_{err}$. The sentences in $S$ are passed as input to the sentence-level VAE. The sampled latent representations from this VAE are used as the representations of the sentences in $S$. These representations are  passed as input to the BLM-level VAE, in the same order as $S$. An instance for the sentence-level VAE consists of a triple $(in, out^+, out^-)$. For our two-level system, we must construct this triple from the input BLM instance: $in \in S$, $out^+ = in$, and $out^- = \{s_k| s_k \in S, s_k \ne in\}$.


\begin{figure}[t]
        \centering
        \includegraphics[width=0.49\textwidth]{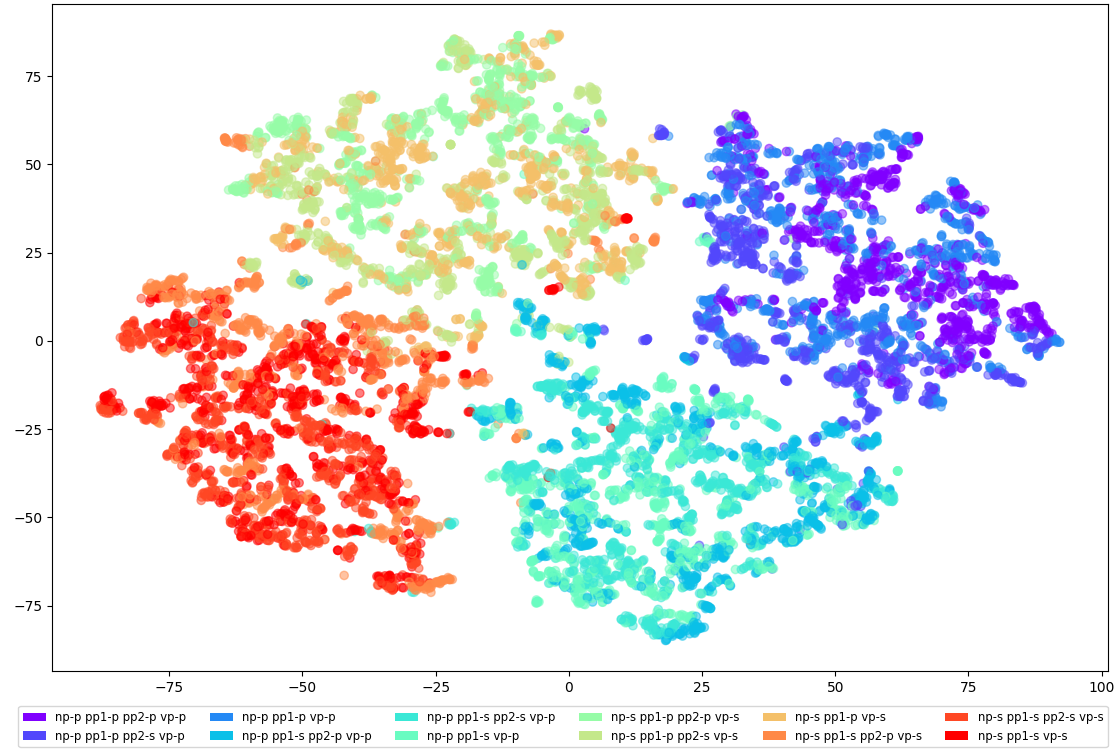}
        \vspace{-5mm}
        \smallpar{TSNE projection of latent representations from the latent layer of the sentence level for the sentences in BLM contexts in the training data, coloured by the chunk pattern.} ~\\          
        
        \includegraphics[width=0.48\textwidth]{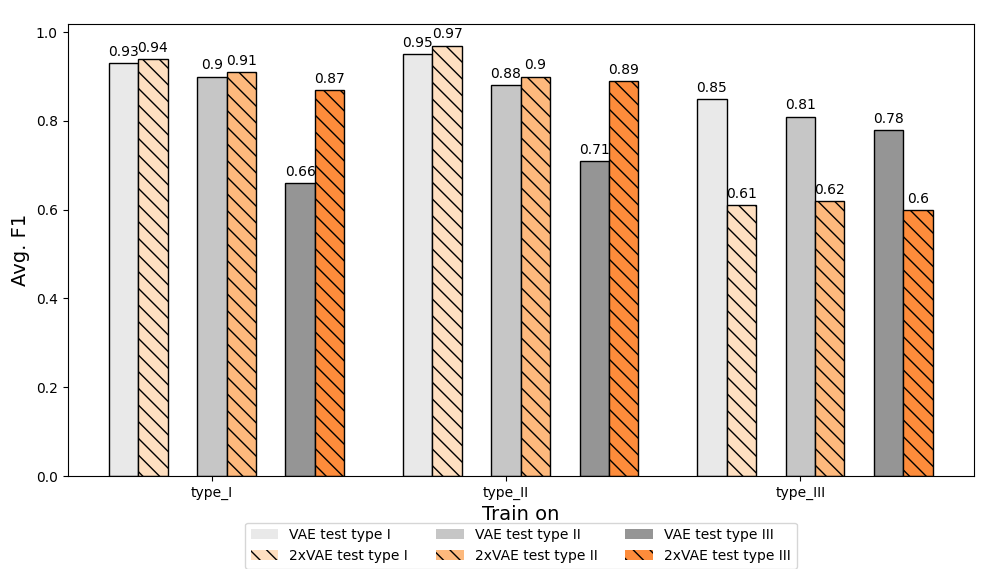}
        
        \smallpar{Average F1 score over 3 runs, grouped by training data on the x-axis, tested on type I, II, III in different shades.} ~\\
        
        \includegraphics[width=0.48\textwidth]{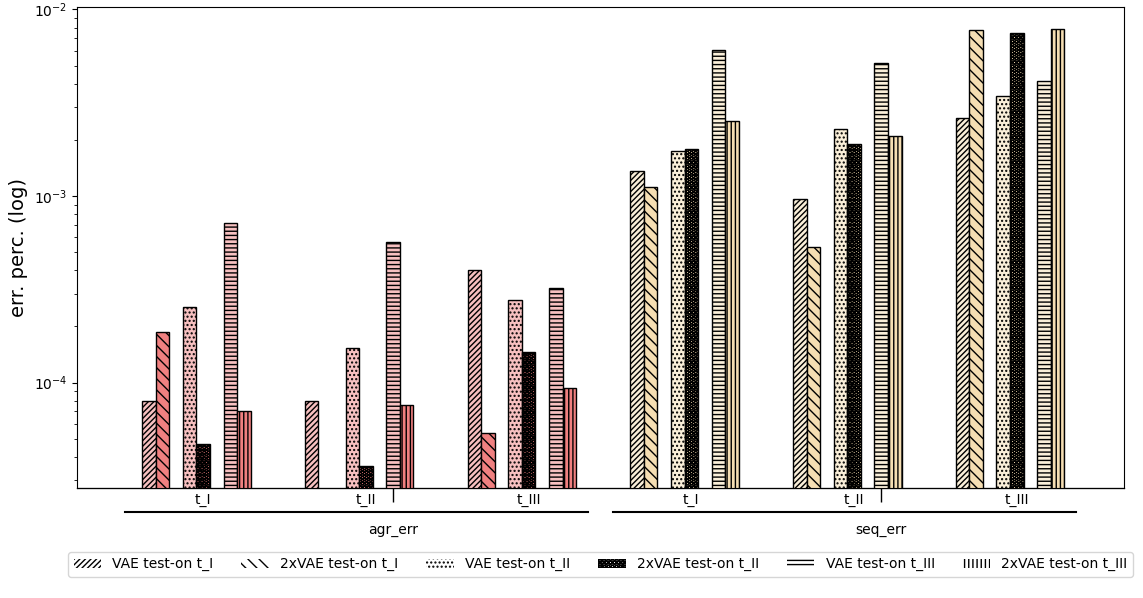}
        \vspace{-5mm}
        \smallpar{Sequence vs. agreement errors analysis.}
    
    \caption{VAE vs 2-level VAE (2xVAE) on the agreement BLM problem}
    \label{fig:agreementF1}
    \vspace{-0.6cm}
\end{figure}

The loss combines the loss signal from the two levels:

\noindent $loss = \\ \mbox{~~~~~~} maxmargin_{sent} + KL_{sent} + \\ \mbox{~~~~~~} maxmargin_{task} + KL_{seq}$

The $maxmargin$ and the scoring of the reconstructed sentence at the sentence level, and the constructed answer at the task level are computed as described in Section \ref{sec:sentences}.

We run experiments on the BLMs for agreement and for verb alternation. While the information necessary to solve the agreement task is more structural, solving the verb alternation task requires not only structural information concerning chunks, but also semantic information, as syntactically similar chunks play different roles in a sentence.

\subsubsection{Analysis}

The results show that this organisation of the system leads to better results compared to the one-level process for these structure-based linguistic problems, thereby providing additional support to our hypothesis that chunks and their attributes are detectable in sentence embeddings. 


We provide results in terms of F1 scores on the task, and analysis of the representations on the latent layer of the sentence level of the system.

Figure \ref{fig:agreementF1} shows the results on the BLM agreement task and the error analysis (detailed results are in the appendix). The results on the task (left panel) provide several insights.
First, from the latent representation analysis, we note that while the sentence representations on the latent layer are not as crisply separated by their chunk pattern as for the experiment in Section \ref{sec:sentences}, there is a clear separation in terms of the grammatical number of the subject and the verb. This is not surprising as the focus of the task is subject-verb agreement. However, as the further results in term of F1 and error analysis on the task show, there is enough information in these compressed latent representation to capture the structural regularities imposed by the patterns of chunks in the input sequence. 

Second, from the results in terms of F1, we note that the two-level process generalizes better from simpler data -- learning on type I and type II leads to better results on all test data, with the highest improvement when tested on type III data, which has the highest lexical variation.

Furthermore, the two-level models learned when training on the lexically simpler data perform better when tested on the type III data than the models learned on type III data itself. This result not only indicates that structure information is more easily detectable when lexical variation is less of a factor, but more importantly, that chunk information is separable from other types of information in the sentence embedding, as the patterns detecting it can be applied successfully for data with additional (lexical) variation.%
%
\footnote{It might appear surprising that the two-level approach leads to lower performance on type III data, particularly when lexical variation had not been an issue for the sentence representation analysis (Section \ref{sec:sentences}). The difference comes from the way the instances were formed, on the fly, for the two-level process: the positive sentence to be reconstructed is the same as the input, instead of being a sentence that has the same structure, but different lexical material. This is because all sentences in the sequence have different structures. We think this weakens the (indirect) supervision signal -- as the correct answer is distinct from the other options. This is not the case for type I and II data, where, because of the very similar lexical material, the distinction between the correct and incorrect answers reduce to the structure. We plan to confirm this in future work using a pre-trained sentence-level VAE.}

\begin{figure}[t]
    \centering

    \includegraphics[width=0.49\textwidth]{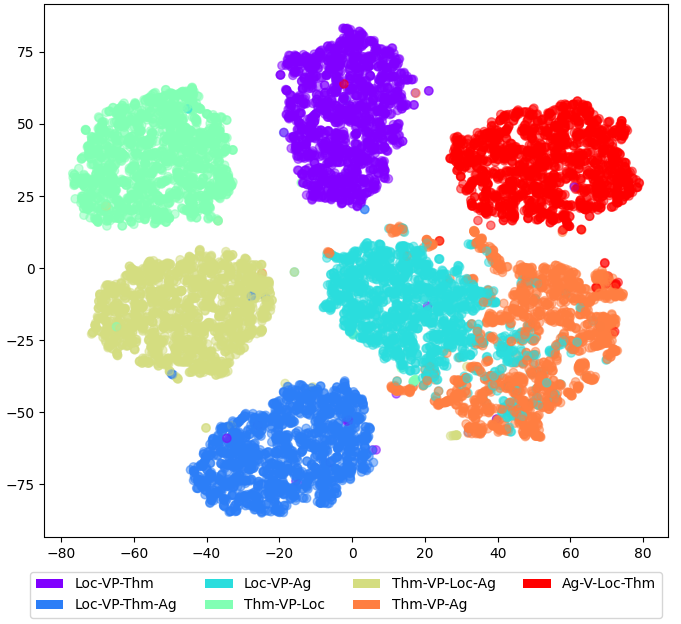}
    \smallpar{TSNE projection of latent representations from the latent layer of the sentence level for the sentences in BLM contexts in the training data, coloured by the pattern of semantic roles.} ~\\
    
    \includegraphics[width=0.48\textwidth]{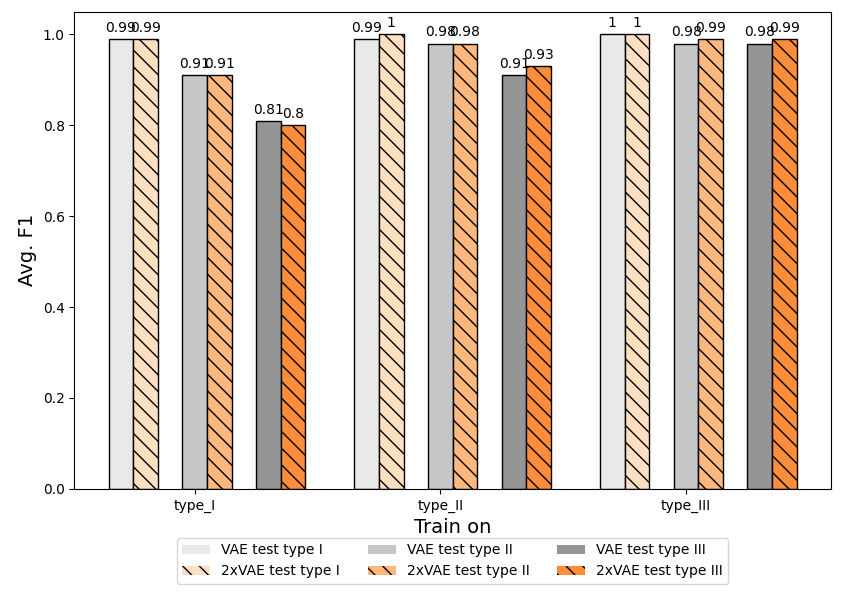}
    
    \smallpar{Average F1 score over 3 runs}
    
    \caption{VAE vs 2-level VAE (2xVAE) on the verb alternation BLM problem, Group 1} 
    \label{fig:verbaltG1}
    
\end{figure}

\begin{figure}[t]
   \centering

    \includegraphics[width=0.49\textwidth]{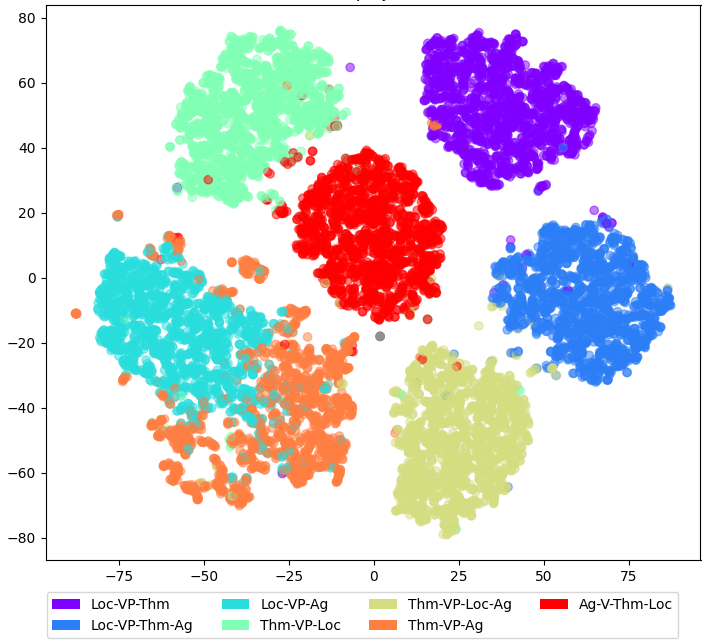}
    \smallpar{TSNE projection of latent representations from the latent layer of the sentence level for the sentences in BLM contexts in the training data, coloured by the pattern of semantic roles.} ~\\
    
    \includegraphics[width=0.5\textwidth]{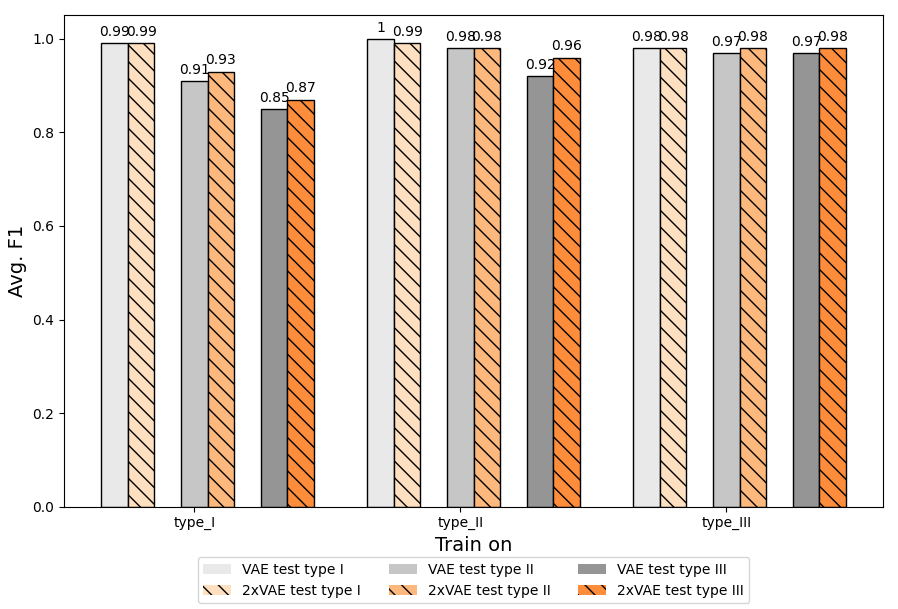}

    \smallpar{Average F1 score over 3 runs}
    
    \caption{VAE vs 2-level VAE (2xVAE) on the verb alternation BLM problem, Group 2} 
    \label{fig:verbaltG2}
    
\end{figure}

Further confirmation of the fact that the sentence level learns to compress sentences into a latent that captures structural information comes from the error analysis, shown in the bottom panel of Figure \ref{fig:agreementF1}. Lower rate of sequence errors, which are correct from the point of view of the targeted phenomenon  -- as described in section \ref{sec:BLMs} -- indicate that there is structure information in the compressed sentence latents.

It is possible that the one-level VAE also detects chunk information in the input sequence, given the high performance on the task. But the fact that the one-level model makes more sequence-based errors indicates that modeling structural information separately  is not only possible, but also beneficial for some tasks.


The results on the verb alternation BLMs are shown in Figures \ref{fig:verbaltG1} and \ref{fig:verbaltG2}. In this problem, and unlike the verb-agreement BLM task, structurally similar chunks - NPs, PPs -- play different semantic roles in the verb alternation data, as shown in Figure \ref{fig:BLM_structure}. 
Other attributes of chunks that are relevant to the current problem -- in this case, semantic roles -- are separated from the sentence embedding whole. This is apparent not only through the F1 results on the task, but also, and maybe more clearly, from the projection of the latent representations from the sentence level, where the separation of the different chunk syntactic and semantic patterns is clear for both groups. For both data subsets, the closest representations are two that have the same syntactic pattern: \textit{NP VerbPass PP}, but semantically differ: \textit{NP-Theme VerbPass PP-Agent} vs. 
\textit{NP-Loc VerbPass PP-Agent}. 

\subsection{Discussion}
We performed two types of experiments: (i) using individual sentences, and an indirect supervision signal about the sentence structure, (ii) incorporating a sentence representation compression step in a task-specific setting. We used two tasks, one which relies on more structural information (subject-verb agreement), and one that also relies on semantic information about the chunks (verb alternation). 

We have investigated each set-up in terms of the results on the task -- as average F1 scores, and through error analysis -- and in terms of internal representations on the latent layer of an encoder-decoder architecture. 

This dual analysis allows us to conclude not only that a task is solved correctly, but that it is solved using structural, morphological and semantic information from the sentence.  We found that information about (varying numbers of) chunks -- noun, verb and prepositional phrases -- and their task-relevant attributes, whether morphological or semantic, can be detected in sentence embeddings from a pretrained transformer model.



\section{Conclusions}




Sentence embeddings obtained from transformer models are compact representations, compressing much knowledge -- morphological, grammatical, semantic, pragmatic --, expressed in text fragments of various length, into a vector of real numbers of fixed length.  If we view the sentence embedding as overlapping layers of information, in a manner similar to audio signals which  consist of overlapping signals of different frequencies, we can distinguish specific information among these layers. In particular, we have shown that we can detect information about chunks -- noun/verb/prepositional phrases -- and their task-relevant attributes in these compact sentence representations. 

These building blocks can be further used in lexically-novel instances to solve tasks that require analytical reasoning, demonstrating that solutions to this task are achieved through abstract  steps typical of fluid intelligence.

\section{Limitations}

We have performed experiments on datasets containing sentences with specific structure and properties to be able to determine whether the type of information we targeted can be detected in sentence embeddings. We applied our framework on a particular pretrained transformer model -- Electra -- which we chose because of the stronger influence of the full context on producing sentence embeddings. Different transformer models may produce different encoding patterns in the sentence embeddings.

\bibliography{custom,anthology}

\begin{thebibliography}{48}
\providecommand{\natexlab}[1]{#1}

\bibitem[{Abney(1992)}]{abney-1992-prosodic}
Steven Abney. 1992.
\newblock \href {https://aclanthology.org/H92-1086} {Prosodic structure,
  performance structure and phrase structure}.
\newblock In \emph{Speech and Natural Language: Proceedings of a Workshop Held
  at Harriman, New York, {F}ebruary 23-26, 1992}.

\bibitem[{An et~al.(2023)An, Jiang, A.~Rodriguez, Nastase, and
  Merlo}]{an-etal-2023-blm}
Aixiu An, Chunyang Jiang, Maria A.~Rodriguez, Vivi Nastase, and Paola Merlo.
  2023.
\newblock \href {https://aclanthology.org/2023.eacl-main.99} {{BLM}-{A}gr{F}: A
  new {F}rench benchmark to investigate generalization of agreement in neural
  networks}.
\newblock In \emph{Proceedings of the 17th Conference of the European Chapter
  of the Association for Computational Linguistics}, pages 1363--1374,
  Dubrovnik, Croatia. Association for Computational Linguistics.

\bibitem[{Bao et~al.(2019)Bao, Zhou, Huang, Li, Mou, Vechtomova, Dai, and
  Chen}]{bao-etal-2019-generating}
Yu~Bao, Hao Zhou, Shujian Huang, Lei Li, Lili Mou, Olga Vechtomova, Xin-yu Dai,
  and Jiajun Chen. 2019.
\newblock \href {https://doi.org/10.18653/v1/P19-1602} {Generating sentences
  from disentangled syntactic and semantic spaces}.
\newblock In \emph{Proceedings of the 57th Annual Meeting of the Association
  for Computational Linguistics}, pages 6008--6019, Florence, Italy.
  Association for Computational Linguistics.

\bibitem[{Belinkov(2022)}]{belinkov-2022-probing}
Yonatan Belinkov. 2022.
\newblock \href {https://doi.org/10.1162/coli_a_00422} {Probing classifiers:
  Promises, shortcomings, and advances}.
\newblock \emph{Computational Linguistics}, 48(1):207--219.

\bibitem[{Bowman et~al.(2016)Bowman, Vilnis, Vinyals, Dai, Jozefowicz, and
  Bengio}]{bowman-etal-2016-generating}
Samuel~R. Bowman, Luke Vilnis, Oriol Vinyals, Andrew Dai, Rafal Jozefowicz, and
  Samy Bengio. 2016.
\newblock \href {https://doi.org/10.18653/v1/K16-1002} {Generating sentences
  from a continuous space}.
\newblock In \emph{Proceedings of the 20th {SIGNLL} Conference on Computational
  Natural Language Learning}, pages 10--21, Berlin, Germany. Association for
  Computational Linguistics.

\bibitem[{Carpenter et~al.(1990)Carpenter, Just, and Shell}]{carpenter1990one}
Patricia~A Carpenter, Marcel~A Just, and Peter Shell. 1990.
\newblock What one intelligence test measures: a theoretical account of the
  processing in the raven progressive matrices test.
\newblock \emph{Psychological review}, 97(3):404.

\bibitem[{Chen et~al.(2019)Chen, Tang, Wiseman, and
  Gimpel}]{chen-etal-2019-multi}
Mingda Chen, Qingming Tang, Sam Wiseman, and Kevin Gimpel. 2019.
\newblock \href {https://doi.org/10.18653/v1/N19-1254} {A multi-task approach
  for disentangling syntax and semantics in sentence representations}.
\newblock In \emph{Proceedings of the 2019 Conference of the North {A}merican
  Chapter of the Association for Computational Linguistics: Human Language
  Technologies, Volume 1 (Long and Short Papers)}, pages 2453--2464,
  Minneapolis, Minnesota. Association for Computational Linguistics.

\bibitem[{Chi et~al.(2020)Chi, Hewitt, and Manning}]{chi-etal-2020-finding}
Ethan~A. Chi, John Hewitt, and Christopher~D. Manning. 2020.
\newblock \href {https://doi.org/10.18653/v1/2020.acl-main.493} {Finding
  universal grammatical relations in multilingual {BERT}}.
\newblock In \emph{Proceedings of the 58th Annual Meeting of the Association
  for Computational Linguistics}, pages 5564--5577, Online. Association for
  Computational Linguistics.

\bibitem[{Clark et~al.(2019)Clark, Khandelwal, Levy, and
  Manning}]{clark-etal-2019-bert}
Kevin Clark, Urvashi Khandelwal, Omer Levy, and Christopher~D. Manning. 2019.
\newblock \href {https://doi.org/10.18653/v1/W19-4828} {What does {BERT} look
  at? an analysis of {BERT}{'}s attention}.
\newblock In \emph{Proceedings of the 2019 ACL Workshop BlackboxNLP: Analyzing
  and Interpreting Neural Networks for NLP}, pages 276--286, Florence, Italy.
  Association for Computational Linguistics.

\bibitem[{Clark et~al.(2020)Clark, Luong, Le, and Manning}]{clark2020electra}
Kevin Clark, Minh-Thang Luong, Quoc~V. Le, and Christopher~D. Manning. 2020.
\newblock \href {https://openreview.net/pdf?id=r1xMH1BtvB} {{ELECTRA}:
  Pre-training text encoders as discriminators rather than generators}.
\newblock In \emph{ICLR}.

\bibitem[{Collins(1997)}]{collins-1997-three}
Michael Collins. 1997.
\newblock \href {https://doi.org/10.3115/976909.979620} {Three generative,
  lexicalised models for statistical parsing}.
\newblock In \emph{35th Annual Meeting of the Association for Computational
  Linguistics and 8th Conference of the {E}uropean Chapter of the Association
  for Computational Linguistics}, pages 16--23, Madrid, Spain. Association for
  Computational Linguistics.

\bibitem[{Conia and Navigli(2022)}]{conia-navigli-2022-probing}
Simone Conia and Roberto Navigli. 2022.
\newblock \href {https://doi.org/10.18653/v1/2022.acl-long.316} {Probing for
  predicate argument structures in pretrained language models}.
\newblock In \emph{Proceedings of the 60th Annual Meeting of the Association
  for Computational Linguistics (Volume 1: Long Papers)}, pages 4622--4632,
  Dublin, Ireland. Association for Computational Linguistics.

\bibitem[{Conneau and Kiela(2018)}]{conneau-kiela-2018-senteval}
Alexis Conneau and Douwe Kiela. 2018.
\newblock \href {https://aclanthology.org/L18-1269} {{S}ent{E}val: An
  evaluation toolkit for universal sentence representations}.
\newblock In \emph{Proceedings of the Eleventh International Conference on
  Language Resources and Evaluation ({LREC} 2018)}, Miyazaki, Japan. European
  Language Resources Association (ELRA).

\bibitem[{Conneau et~al.(2018)Conneau, Kruszewski, Lample, Barrault, and
  Baroni}]{conneau2018}
Alexis Conneau, German Kruszewski, Guillaume Lample, Lo{\"\i}c Barrault, and
  Marco Baroni. 2018.
\newblock \href {https://doi.org/10.18653/v1/P18-1198} {What you can cram into
  a single {\$}{\&}!{\#}* vector: Probing sentence embeddings for linguistic
  properties}.
\newblock In \emph{Proceedings of the 56th Annual Meeting of the Association
  for Computational Linguistics (Volume 1: Long Papers)}, pages 2126--2136,
  Melbourne, Australia. Association for Computational Linguistics.

\bibitem[{de~Vries et~al.(2020)de~Vries, van Cranenburgh, and
  Nissim}]{de-vries-etal-2020-whats}
Wietse de~Vries, Andreas van Cranenburgh, and Malvina Nissim. 2020.
\newblock \href {https://doi.org/10.18653/v1/2020.findings-emnlp.389} {What{'}s
  so special about {BERT}{'}s layers? a closer look at the {NLP} pipeline in
  monolingual and multilingual models}.
\newblock In \emph{Findings of the Association for Computational Linguistics:
  EMNLP 2020}, pages 4339--4350, Online. Association for Computational
  Linguistics.

\bibitem[{Devlin et~al.(2018)Devlin, Chang, Lee, and
  Toutanova}]{devlin-etal-2018-bert}
Jacob Devlin, Ming{-}Wei Chang, Kenton Lee, and Kristina Toutanova. 2018.
\newblock \href {https://arxiv.org/abs/1810.04805} {{BERT:} pre-training of
  deep bidirectional transformers for language understanding}.
\newblock \emph{CoRR}, abs/1810.04805.

\bibitem[{Devlin et~al.(2019)Devlin, Chang, Lee, and
  Toutanova}]{devlin-etal-2019-bert}
Jacob Devlin, Ming-Wei Chang, Kenton Lee, and Kristina Toutanova. 2019.
\newblock \href {https://doi.org/10.18653/v1/N19-1423} {{BERT}: Pre-training of
  deep bidirectional transformers for language understanding}.
\newblock In \emph{Proceedings of the 2019 Conference of the North {A}merican
  Chapter of the Association for Computational Linguistics: Human Language
  Technologies, Volume 1 (Long and Short Papers)}, pages 4171--4186,
  Minneapolis, Minnesota. Association for Computational Linguistics.

\bibitem[{Elazar et~al.(2021)Elazar, Ravfogel, Jacovi, and
  Goldberg}]{elazar-etal-2021-amnesic}
Yanai Elazar, Shauli Ravfogel, Alon Jacovi, and Yoav Goldberg. 2021.
\newblock \href {https://doi.org/10.1162/tacl_a_00359} {Amnesic probing:
  Behavioral explanation with amnesic counterfactuals}.
\newblock \emph{Transactions of the Association for Computational Linguistics},
  9:160--175.

\bibitem[{Franck et~al.(2002)Franck, Vigliocco, and Nicol}]{franck2002subject}
Julie Franck, Gabriella Vigliocco, and Janet. Nicol. 2002.
\newblock Subject-verb agreement errors in {F}rench and {E}nglish: The role of
  syntactic hierarchy.
\newblock \emph{Language and Cognitive Processes}, 17(4):371–404.

\bibitem[{Hewitt and Manning(2019)}]{hewitt-manning-2019-structural}
John Hewitt and Christopher~D. Manning. 2019.
\newblock \href {https://doi.org/10.18653/v1/N19-1419} {{A} structural probe
  for finding syntax in word representations}.
\newblock In \emph{Proceedings of the 2019 Conference of the North {A}merican
  Chapter of the Association for Computational Linguistics: Human Language
  Technologies, Volume 1 (Long and Short Papers)}, pages 4129--4138,
  Minneapolis, Minnesota. Association for Computational Linguistics.

\bibitem[{Htut et~al.(2019)Htut, Phang, Bordia, and Bowman}]{htut2019attention}
Phu~Mon Htut, Jason Phang, Shikha Bordia, and Samuel~R. Bowman. 2019.
\newblock \href {https://arxiv.org/abs/1911.12246} {Do attention heads in bert
  track syntactic dependencies?}
\newblock \emph{Preprint}, arXiv:1911.12246.

\bibitem[{Ippolito et~al.(2020)Ippolito, Grangier, Eck, and
  Callison-Burch}]{ippolito-etal-2020-toward}
Daphne Ippolito, David Grangier, Douglas Eck, and Chris Callison-Burch. 2020.
\newblock \href {https://doi.org/10.18653/v1/2020.acl-main.666} {Toward better
  storylines with sentence-level language models}.
\newblock In \emph{Proceedings of the 58th Annual Meeting of the Association
  for Computational Linguistics}, pages 7472--7478, Online. Association for
  Computational Linguistics.

\bibitem[{Jawahar et~al.(2019)Jawahar, Sagot, and
  Seddah}]{jawahar-etal-2019-bert}
Ganesh Jawahar, Beno{\^\i}t Sagot, and Djam{\'e} Seddah. 2019.
\newblock \href {https://doi.org/10.18653/v1/P19-1356} {What does {BERT} learn
  about the structure of language?}
\newblock In \emph{Proceedings of the 57th Annual Meeting of the Association
  for Computational Linguistics}, pages 3651--3657, Florence, Italy.
  Association for Computational Linguistics.

\bibitem[{Kingma and Welling(2013)}]{kingma2013vae}
Diederik~P Kingma and Max Welling. 2013.
\newblock Auto-encoding variational bayes.
\newblock \emph{arXiv preprint arXiv:1312.6114}.

\bibitem[{Lan et~al.(2019)Lan, Chen, Goodman, Gimpel, Sharma, and
  Soricut}]{lan-etal-2020-albert}
Zhenzhong Lan, Mingda Chen, Sebastian Goodman, Kevin Gimpel, Piyush Sharma, and
  Radu Soricut. 2019.
\newblock \href {https://arxiv.org/abs/1909.11942} {{ALBERT:} {A} lite {BERT}
  for self-supervised learning of language representations}.
\newblock \emph{CoRR}, abs/1909.11942.

\bibitem[{Liu et~al.(2019)Liu, Ott, Goyal, Du, Joshi, Chen, Levy, Lewis,
  Zettlemoyer, and Stoyanov}]{liu2019roberta}
Yinhan Liu, Myle Ott, Naman Goyal, Jingfei Du, Mandar Joshi, Danqi Chen, Omer
  Levy, Mike Lewis, Luke Zettlemoyer, and Veselin Stoyanov. 2019.
\newblock Roberta: A robustly optimized bert pretraining approach.
\newblock \emph{arXiv preprint arXiv:1907.11692}.

\bibitem[{Luo(2021)}]{luo-2021-attention}
Ziyang Luo. 2021.
\newblock \href {https://doi.org/10.18653/v1/2021.eacl-srw.2} {Have attention
  heads in {BERT} learned constituency grammar?}
\newblock In \emph{Proceedings of the 16th Conference of the European Chapter
  of the Association for Computational Linguistics: Student Research Workshop},
  pages 8--15, Online. Association for Computational Linguistics.

\bibitem[{Mercatali and
  Freitas(2021)}]{mercatali-freitas-2021-disentangling-generative}
Giangiacomo Mercatali and Andr{\'e} Freitas. 2021.
\newblock \href {https://doi.org/10.18653/v1/2021.findings-emnlp.301}
  {Disentangling generative factors in natural language with discrete
  variational autoencoders}.
\newblock In \emph{Findings of the Association for Computational Linguistics:
  EMNLP 2021}, pages 3547--3556, Punta Cana, Dominican Republic. Association
  for Computational Linguistics.

\bibitem[{Merlo(2023)}]{merlo2023}
Paola Merlo. 2023.
\newblock \href {https://doi.org/10.48550/arXiv.2306.11444} {Blackbird language
  matrices {(BLM)}, a new task for rule-like generalization in neural networks:
  Motivations and formal specifications}.
\newblock \emph{ArXiv}, cs.CL 2306.11444.

\bibitem[{Nastase and Merlo(2023)}]{nastase-merlo-2023-grammatical}
Vivi Nastase and Paola Merlo. 2023.
\newblock \href {https://doi.org/10.18653/v1/2023.repl4nlp-1.3} {Grammatical
  information in {BERT} sentence embeddings as two-dimensional arrays}.
\newblock In \emph{Proceedings of the 8th Workshop on Representation Learning
  for NLP (RepL4NLP 2023)}, pages 22--39, Toronto, Canada. Association for
  Computational Linguistics.

\bibitem[{Nikolaev and
  Pad{\'o}(2023{\natexlab{a}})}]{nikolaev-pado-2023-investigating}
Dmitry Nikolaev and Sebastian Pad{\'o}. 2023{\natexlab{a}}.
\newblock \href {https://doi.org/10.18653/v1/2023.blackboxnlp-1.11}
  {Investigating semantic subspaces of transformer sentence embeddings through
  linear structural probing}.
\newblock In \emph{Proceedings of the 6th BlackboxNLP Workshop: Analyzing and
  Interpreting Neural Networks for NLP}, pages 142--154, Singapore. Association
  for Computational Linguistics.

\bibitem[{Nikolaev and
  Pad{\'o}(2023{\natexlab{b}})}]{nikolaev-pado-2023-representation}
Dmitry Nikolaev and Sebastian Pad{\'o}. 2023{\natexlab{b}}.
\newblock \href {https://aclanthology.org/2023.eacl-main.268} {Representation
  biases in sentence transformers}.
\newblock In \emph{Proceedings of the 17th Conference of the European Chapter
  of the Association for Computational Linguistics}, pages 3701--3716,
  Dubrovnik, Croatia. Association for Computational Linguistics.

\bibitem[{Nikolaev and
  Pad{\'o}(2023{\natexlab{c}})}]{nikolaev-pado-2023-universe}
Dmitry Nikolaev and Sebastian Pad{\'o}. 2023{\natexlab{c}}.
\newblock \href {https://aclanthology.org/2023.iwcs-1.12} {The universe of
  utterances according to {BERT}}.
\newblock In \emph{Proceedings of the 15th International Conference on
  Computational Semantics}, pages 99--105, Nancy, France. Association for
  Computational Linguistics.

\bibitem[{Niu et~al.(2022)Niu, Lu, and Penn}]{niu-etal-2022-bert}
Jingcheng Niu, Wenjie Lu, and Gerald Penn. 2022.
\newblock \href {https://aclanthology.org/2022.coling-1.278} {Does {BERT}
  rediscover a classical {NLP} pipeline?}
\newblock In \emph{Proceedings of the 29th International Conference on
  Computational Linguistics}, pages 3143--3153, Gyeongju, Republic of Korea.
  International Committee on Computational Linguistics.

\bibitem[{Opitz and Frank(2022)}]{opitz-frank-2022-sbert}
Juri Opitz and Anette Frank. 2022.
\newblock \href {https://aclanthology.org/2022.aacl-main.48} {{SBERT} studies
  meaning representations: Decomposing sentence embeddings into explainable
  semantic features}.
\newblock In \emph{Proceedings of the 2nd Conference of the Asia-Pacific
  Chapter of the Association for Computational Linguistics and the 12th
  International Joint Conference on Natural Language Processing (Volume 1: Long
  Papers)}, pages 625--638, Online only. Association for Computational
  Linguistics.

\bibitem[{Papadimitriou et~al.(2021)Papadimitriou, Chi, Futrell, and
  Mahowald}]{papadimitriou-etal-2021-deep}
Isabel Papadimitriou, Ethan~A. Chi, Richard Futrell, and Kyle Mahowald. 2021.
\newblock \href {https://doi.org/10.18653/v1/2021.eacl-main.215} {Deep
  subjecthood: Higher-order grammatical features in multilingual {BERT}}.
\newblock In \emph{Proceedings of the 16th Conference of the European Chapter
  of the Association for Computational Linguistics: Main Volume}, pages
  2522--2532, Online. Association for Computational Linguistics.

\bibitem[{Rajpurkar et~al.(2016)Rajpurkar, Zhang, Lopyrev, and
  Liang}]{rajpurkar-etal-2016-squad}
Pranav Rajpurkar, Jian Zhang, Konstantin Lopyrev, and Percy Liang. 2016.
\newblock \href {https://doi.org/10.18653/v1/D16-1264} {{SQ}u{AD}: 100,000+
  questions for machine comprehension of text}.
\newblock In \emph{Proceedings of the 2016 Conference on Empirical Methods in
  Natural Language Processing}, pages 2383--2392, Austin, Texas. Association
  for Computational Linguistics.

\bibitem[{Reimers and Gurevych(2019)}]{reimers-gurevych-2019-sentence}
Nils Reimers and Iryna Gurevych. 2019.
\newblock \href {https://doi.org/10.18653/v1/D19-1410} {Sentence-{BERT}:
  Sentence embeddings using {S}iamese {BERT}-networks}.
\newblock In \emph{Proceedings of the 2019 Conference on Empirical Methods in
  Natural Language Processing and the 9th International Joint Conference on
  Natural Language Processing (EMNLP-IJCNLP)}, pages 3982--3992, Hong Kong,
  China. Association for Computational Linguistics.

\bibitem[{Rogers et~al.(2020)Rogers, Kovaleva, and
  Rumshisky}]{rogers-etal-2020-primer}
Anna Rogers, Olga Kovaleva, and Anna Rumshisky. 2020.
\newblock \href {https://doi.org/10.1162/tacl_a_00349} {A primer in
  {BERT}ology: What we know about how {BERT} works}.
\newblock \emph{Transactions of the Association for Computational Linguistics},
  8:842--866.

\bibitem[{Samo et~al.(2023)Samo, Nastase, Jiang, and Merlo}]{samo-etal-2023}
Giuseppe Samo, Vivi Nastase, Chunyang Jiang, and Paola Merlo. 2023.
\newblock {BLM-s/lE}: A structured dataset of {English} spray-load verb
  alternations for testing generalization in {LLM}s.
\newblock In \emph{Findings of the 2023 Conference on Empirical Methods in
  Natural Language Processing}.

\bibitem[{Silva De~Carvalho et~al.(2023)Silva De~Carvalho, Mercatali, Zhang,
  and Freitas}]{carvalho-etal-2023-learning}
Danilo Silva De~Carvalho, Giangiacomo Mercatali, Yingji Zhang, and Andr{\'e}
  Freitas. 2023.
\newblock \href {https://doi.org/10.18653/v1/2023.findings-eacl.101} {Learning
  disentangled representations for natural language definitions}.
\newblock In \emph{Findings of the Association for Computational Linguistics:
  EACL 2023}, pages 1371--1384, Dubrovnik, Croatia. Association for
  Computational Linguistics.

\bibitem[{Strubell et~al.(2018)Strubell, Verga, Andor, Weiss, and
  McCallum}]{strubell-etal-2018-linguistically}
Emma Strubell, Patrick Verga, Daniel Andor, David Weiss, and Andrew McCallum.
  2018.
\newblock \href {https://doi.org/10.18653/v1/D18-1548} {Linguistically-informed
  self-attention for semantic role labeling}.
\newblock In \emph{Proceedings of the 2018 Conference on Empirical Methods in
  Natural Language Processing}, pages 5027--5038, Brussels, Belgium.
  Association for Computational Linguistics.

\bibitem[{Tan et~al.(2018)Tan, Wang, Xie, Chen, and Shi}]{tan2018deep}
Zhixing Tan, Mingxuan Wang, Jun Xie, Yidong Chen, and Xiaodong Shi. 2018.
\newblock Deep semantic role labeling with self-attention.
\newblock In \emph{Proceedings of the AAAI conference on artificial
  intelligence}, volume~32.

\bibitem[{Tenney et~al.(2019{\natexlab{a}})Tenney, Das, and
  Pavlick}]{tenney-etal-2019-bert}
Ian Tenney, Dipanjan Das, and Ellie Pavlick. 2019{\natexlab{a}}.
\newblock \href {https://doi.org/10.18653/v1/P19-1452} {{BERT} rediscovers the
  classical {NLP} pipeline}.
\newblock In \emph{Proceedings of the 57th Annual Meeting of the Association
  for Computational Linguistics}, pages 4593--4601, Florence, Italy.
  Association for Computational Linguistics.

\bibitem[{Tenney et~al.(2019{\natexlab{b}})Tenney, Xia, Chen, Wang, Poliak,
  McCoy, Kim, Van~Durme, Bowman, Das
  et~al.}]{tenney-etal-2019-learn-from-context}
Ian Tenney, Patrick Xia, Berlin Chen, Alex Wang, Adam Poliak, R~Thomas McCoy,
  Najoung Kim, Benjamin Van~Durme, Samuel~R Bowman, Dipanjan Das, et~al.
  2019{\natexlab{b}}.
\newblock What do you learn from context? probing for sentence structure in
  contextualized word representations.
\newblock In \emph{The Seventh International Conference on Learning
  Representations (ICLR)}, pages 235--249.

\bibitem[{Tjong Kim~Sang and
  Buchholz(2000)}]{tjong-kim-sang-buchholz-2000-introduction}
Erik~F. Tjong Kim~Sang and Sabine Buchholz. 2000.
\newblock \href {https://aclanthology.org/W00-0726} {Introduction to the
  {C}o{NLL}-2000 shared task chunking}.
\newblock In \emph{Fourth Conference on Computational Natural Language Learning
  and the Second Learning Language in Logic Workshop}.

\bibitem[{Wang et~al.(2018)Wang, Singh, Michael, Hill, Levy, and
  Bowman}]{wang-etal-2018-glue}
Alex Wang, Amanpreet Singh, Julian Michael, Felix Hill, Omer Levy, and Samuel
  Bowman. 2018.
\newblock \href {https://doi.org/10.18653/v1/W18-5446} {{GLUE}: A multi-task
  benchmark and analysis platform for natural language understanding}.
\newblock In \emph{Proceedings of the 2018 {EMNLP} Workshop {B}lackbox{NLP}:
  Analyzing and Interpreting Neural Networks for {NLP}}, pages 353--355,
  Brussels, Belgium. Association for Computational Linguistics.

\bibitem[{Yi et~al.(2022)Yi, Bruno, Han, Zukerman, and
  Steinert-Threlkeld}]{yi-etal-2022-probing}
David Yi, James Bruno, Jiayu Han, Peter Zukerman, and Shane Steinert-Threlkeld.
  2022.
\newblock \href {https://aclanthology.org/2022.blackboxnlp-1.12} {Probing for
  understanding of {E}nglish verb classes and alternations in large pre-trained
  language models}.
\newblock In \emph{Proceedings of the Fifth BlackboxNLP Workshop on Analyzing
  and Interpreting Neural Networks for NLP}, pages 142--152, Abu Dhabi, United
  Arab Emirates (Hybrid). Association for Computational Linguistics.

\end{thebibliography}

\newpage

\onecolumn
\appendix

\section{Appendix}
\label{sec:appendix}

\subsection{Sentence data}
\label{sec:sentencedata}

To build the sentence data, we use a seed file that was used to generate the subject-verb agreement data. A seed, consisting of noun, prepositional and verb phrases with different grammatical numbers, can be combined to build sentences consisting of different sequences of such chunks. Table \ref{tab:data_samples} includes a partial line from the seed file, from which individual sentences and a BLM instance can be constructed. We use French and English versions of the seed file to build the corresponding datasets.

\begin{table}[h]
  \footnotesize
  \renewcommand{\tabcolsep}{1mm}
    \centering
    \begin{tabular}{p{\x}p{\x}p{\x}p{\y}p{\y}p{\y}p{\y}p{\y}} \hline
    Subj\_sg	& Subj\_pl & P1\_sg	 & P1\_pl & P2\_sg	& P2\_pl &	V\_sg &	V\_pl \\ \hline
   The computer & The computers & with the program	& with the programs &	of the experiment & of the experiments & is broken & are broken	\\ \hline \hline
    \multicolumn{3}{l}{
      \begingroup
       \renewcommand{\arraystretch}{1.5}
       \begin{tabular}{p{3.7cm}p{0.8cm}} \\
       \multicolumn{2}{l}{\bf Sent. with different chunks} \\ \hline
        The computer is broken. & np-s vp-s \\
        The computers are broken. & np-p vp-p \\
        The computer with the program is broken. & np-s pp1-s vp-s \\
        ... & ... \\
        The computers with the programs of the experiments are broken. & np-p pp1-p pp2-p vp-p \\ \hline \vspace{2mm}
       \end{tabular}
       \endgroup
       } 
    & \multicolumn{5}{r}{
     \begingroup
     \renewcommand{\arraystretch}{1.2}
      \begin{tabular}{p{8.5cm}}
         {\bf a BLM instance} \\ \hline
         \ul{Context:} \\
         The computer with the program is broken. \\
         The computers with the program are broken. \\
         The computer with the programs is broken. \\
         The computers with the programs are broken. \\
         The computer with the program of the experiment is broken. \\
         The computers with the program of the experiment are broken. \\
         The computer with the programs of the experiment is broken. \\ \hline
         \ul{Answer set:} \\
         {\it The computers with the programs of the experiment are broken.} \\
        The computers with the programs of the experiments are broken. \\
        The computers with the program of the experiment are broken. \\
        The computers with the program of the experiment is broken. \\
        ... \\ \hline \vspace{2mm}
      \end{tabular}
    \endgroup
    } \\ \hline
    \end{tabular}
    \caption{A line from the seed file on top, and a set of individual sentences built from it, as well as one BLM instance.}
    \label{tab:data_samples}
\end{table}

The algorithm to produce a dataset from the generated sentences is detailed in Figure \ref{tab:data_instances} below.

\begin{figure}[h]
    \small
    \centering
    \begin{minipage}{\textwidth}    
    \begin{algorithmic}
        \State Data = []; $N_{negs}$
        \For{patterns $p$}
        \For{$s_i \in S_p$}
            \State in = $s_i$
            \For{$s_j \in S_p$}
                \State out$^+$ = $s_j$ 
                \State out$^-$ = $\{s_k, k \in range(N_{negs}), s_k \in S_{\neg p} \}$  
                \State Data = Data $\cup$ [(in, out$^+$, out$^-$)]
            \EndFor
          \EndFor
        \EndFor
    \end{algorithmic}
    \end{minipage}
    \caption{Data generation algorithm}
    \label{tab:data_instances}
\end{figure}

\newpage

\subsection{Example of data for the agreement BLM}
\label{app:agr}

\begin{table*}[h!]
{\small
    \centering
    \begin{tabular}{l}\hline
    {\bf Example subject NPs} from \cite{franck2002subject} \\ \hline
    {\it L'ordinateur avec le programme de l'experience} \\
    The computer with the program of the experiments
    \\ \hline \hline
    {\bf Manually expanded and completed sentences} \\ \hline
    {\it L'ordinateur avec le programme de l'experience est en panne.}\\
    The computer with the program of the experiments is down. \\
    {\it Jean suppose que l'ordinateur avec le programme de l'experience est en panne.} \\
    Jean thinks that the computer with the program of the experiments is down. \\
    {\it L'ordinateur avec le programme dont Jean se servait est en panne.} \\
    The computer with the program that John was using is down.
    \\ \hline 
    \end{tabular} \\ ~\\ 
    \begin{tabularx}{\textwidth}{l|l|l|l|l|l}\hline
    \multicolumn{6}{l}{\bf A seed for language matrix generation} \\ \hline
    {\it Jean suppose que} & {\it l'ordinateur} & {\it avec le programme} & {\it de l'experience} & {\it dont Jean se servait} & {\it est en panne} \\
    Jean thinks that & the computer & with the program & of the experiment & that John was using & is down \\
    & {\it les ordinateurs} & {\it avec les programmes} & & & {\it sont en panne} \\
    & the computers & with the programs & & & are down 
    \\ \hline
    \end{tabularx}
    }
    \caption{Examples from \cite{franck2002subject}, manually completed and expanded sentences based on these examples, and seeds made based on these sentences for subject-verb agreement BLM dataset that contain all number variations for the nouns and the verb.}
    \label{tab:franckexamples}
\end{table*}

\begin{table*}[h!]

\small
{
\begin{tabularx}{\textwidth}{rllllll} 
\hline
\multicolumn{7}{l}{Main clause}                                                                                                                                                               \\ 
\hline
1 &                  & \textcolor{blue}{L'ordinateur}   & \textcolor{blue}{avec le programme}  &                                    &                      & \textcolor{blue}{est en panne.}  \\
2 &                  & \textcolor{red}{Les ordinateurs} & \textcolor{blue}{avec le programme}  &                                    &                      & \textcolor{red}{sont en panne.}  \\
3 &                  & \textcolor{blue}{L'ordinateur}   & \textcolor{red}{avec les programmes} &                                    &                      & \textcolor{blue}{est en panne.}  \\
4 &                  & \textcolor{red}{Les ordinateurs} & \textcolor{red}{avec les programmes} &                                    &                      & \textcolor{red}{sont en panne.}  \\
5 &                  & \textcolor{blue}{L'ordinateur}   & \textcolor{blue}{avec le programme}  & \textcolor{blue}{de l'exp\'{e}rience} &                      & \textcolor{blue}{est en panne.}  \\
6 &                  & \textcolor{red}{Les ordinateurs} & \textcolor{blue}{avec le programme}  & \textcolor{blue}{de l'exp\'{e}rience} &                      & \textcolor{red}{sont en panne.}  \\
7 &                  & \textcolor{blue}{L'ordinateur}   & \textcolor{red}{avec les programmes} & \textcolor{blue}{de l'exp\'{e}rience} &                      & \textcolor{blue}{est en panne.}  \\
8 &                  & Les \textcolor{red}{ordinateurs} & \textcolor{red}{avec les programmes} & \textcolor{blue}{de l'exp\'{e}rience} &                      & \textcolor{red}{sont en panne.}  \\ 
\hline\hline
\multicolumn{7}{l}{Completive clause}                                                                                                                                                         \\ 
\hline
1 & Jean suppose que & \textcolor{blue}{l'ordinateur}   & \textcolor{blue}{avec le programme}  &                                    &                      & \textcolor{blue}{est en panne.}  \\
2 & Jean suppose que & \textcolor{red}{les ordinateurs} & \textcolor{blue}{avec le programme}  &                                    &                      & \textcolor{red}{sont en panne.}  \\
3 & Jean suppose que & \textcolor{blue}{l'ordinateur}   & \textcolor{red}{avec les programmes} &                                    &                      & \textcolor{blue}{est en panne.}  \\
4 & Jean suppose que & \textcolor{red}{les ordinateurs} & \textcolor{red}{avec les programmes} &                                    &                      & \textcolor{red}{sont en panne.}  \\
5 & Jean suppose que & \textcolor{blue}{l'ordinateur}   & \textcolor{blue}{avec le programme}  & \textcolor{blue}{de l'exp\'{e}rience} &                      & \textcolor{blue}{est en panne.}  \\
6 & Jean suppose que & \textcolor{red}{les ordinateurs} & \textcolor{blue}{avec le programme}  & \textcolor{blue}{de l'exp\'{e}rience} &                      & \textcolor{red}{sont en panne.}  \\
7 & Jean suppose que & \textcolor{blue}{l'ordinateur}   & \textcolor{red}{avec les programmes} & \textcolor{blue}{de l'exp\'{e}rience} &                      & \textcolor{blue}{est en panne.}  \\
8 & Jean suppose que & \textcolor{red}{les ordinateurs} & \textcolor{red}{avec les programmes} & \textcolor{blue}{de l'exp\'{e}rience} &                      & \textcolor{red}{sont en panne.}  \\ 
\hline\hline
\multicolumn{7}{l}{Relative clause}                                                                                                                                                           \\ 
\hline
1 &                  & \textcolor{blue}{L'ordinateur}   & \textcolor{blue}{avec le programme}  &                                    & dont Jean se servait & \textcolor{blue}{est en panne.}  \\
2 &                  & \textcolor{red}{Les ordinateurs} & \textcolor{blue}{avec le programme}  &                                    & dont Jean se servait & \textcolor{red}{sont en panne.}  \\
3 &                  & \textcolor{blue}{L'ordinateur}   & \textcolor{red}{avec les programmes} &                                    & dont Jean se servait & \textcolor{blue}{est en panne.}  \\
4 &                  & \textcolor{red}{Les ordinateurs} & \textcolor{red}{avec les programmes} &                                    & dont Jean se servait & \textcolor{red}{sont en panne.}  \\
5 &                  & \textcolor{blue}{L'ordinateur}   & \textcolor{blue}{avec le programme}  & \textcolor{blue}{de l'exp\'{e}rience} & dont Jean se servait & \textcolor{blue}{est en panne.}  \\
6 &                  & \textcolor{red}{Les ordinateurs} & \textcolor{blue}{avec le programme}  & \textcolor{blue}{de l'exp\'{e}rience} & dont Jean se servait & \textcolor{red}{sont en panne.}  \\
7 &                  & \textcolor{blue}{L'ordinateur}   & \textcolor{red}{avec les programmes} & \textcolor{blue}{de l'exp\'{e}rience} & dont Jean se servait & \textcolor{blue}{est en panne.}  \\
8 &                  & \textcolor{red}{Les ordinateurs} & \textcolor{red}{avec les programmes} & \textcolor{blue}{de l'exp\'{e}rience} & dont Jean se servait & \textcolor{red}{sont en panne.}  \\
\hline
\end{tabularx}
 \\ \\ \\
  \begin{tabular}{rll} \hline
 \multicolumn{3}{c}{{\bf Answer set} for problem constructed from lines 1-7 of the main clause sequence} \\ \hline
 1 & L'ordinateur avec le programme et l'experi\'{e}nce est en panne. & N2 coord N3 \\ 
 2 & {\bf Les ordinateurs avec les programmes de l'experi\'{e}nce sont en panne.} & correct \\
 3 & L'ordinateur avec le programme est en panne. & wrong number of attractors \\
 4 & L'ordinateur avec le program de l'experi\'{e}nce sont en panne. & agreement error \\
 5 & Les ordinateurs avec les programmes de l'experi\'{e}nce sont en panne. & wrong nr. for 1$^{st}$ attractor noun \\
 6 & Les ordinateurs avec les programmes de les experi\'{e}nces sont en panne. & wrong nr. for 2$^{nd}$ attractor noun \\ \hline
 \end{tabular}

}
\caption{BLM instances for verb-subject agreement, with 2 attractors (programme, experi\'{e}nce), and three clause structures. And candidate answer set for a problem constructed from lines 1-7 of the main clause sequence.}
\label{tab:matrices}
\end{table*}

\newpage

\subsection{Example of data for the verb alternation BLM}
\label{app:verbalt}

\begin{figure}[h!]
  \small
  \setlength{\tabcolsep}{1mm}
    \begin{tabular}{llll} 
    \textsc{Type I} \\       
    \\\hline 
    \textsc{Example of  context}\\ \hline 
    	The buyer can load the tools in bags.\\ 
        The tools were loaded by the buyer \\
        The tools were loaded in bags by the buyer\\
        The tools were loaded in bags\\
        Bags were loaded by the buyer \\
        Bags were loaded with the tools by the buyer \\
        Bags were loaded with the tools \\
    	??? \\ 
        \\ \hline 
    
        \sc Example of  answers\\ \hline 
        \textbf{The buyer can load bags with the tools}\\
        The buyer was loaded bags with the tools \\
        The buyer can load bags the tools\\
        The buyer can load in bags with the tools \\
        The buyer can load bags on sale\\
        The buyer can load bags under the tools\\
        Bags can load the buyer with the tools\\
        The tools can load the buyer in bags\\
        Bags can load the tools in the buyer\\
    \hline 
    \end{tabular}
         \caption{Example of Type I context sentences and answer set.}
       \label{BLMs/lE-typeicompl}
\end{figure}

\subsection{Experimental details}

All systems used a learning rate of 0.001 and Adam optimizer, and batch size 100. The system was trained for 300 epochs for all experiments.

The experiments were run on an HP PAIR Workstation Z4 G4 MT, with an Intel Xeon W-2255 processor, 64G RAM, and a MSI GeForce RTX 3090 VENTUS 3X OC 24G GDDR6X GPU.

The {\bf sentence-level encoder decoder} has 106 603 parameters. It consists of an encoder with a CNN layer followed by a FFNN layer. The CNN input has shape 32x24. We use a kernel size 15x15 with stride 1x1, and 40 channels. The linearized CNN output has 240 units, which the FFNN compresses into the latent layer of size 5+5 (mean+std). The decoder is a mirror of the encoder, which expands a sampled latent of size 5 into a 32x24 representation.

The {\bf two-level system} consists of the sentence level encoder-decoder described above, and a task-specific layer. The input to the task layer is a 7x5 input (sequence of 7 sentences, whose representation we obtain from the latent of the sentence level), which is compressed using a CNN with kernel 4x4 and stride 1x1 and 32 channels into ... units, which are compressed using a FFNN layer into a latent layer of size 5+5 (mean+std). The decoder consists of a FFNN which expands the sampled latent of size 5 into 7200 units, which are then processed through a CNN with kernel size 15x15 and stride 1x1, and produces a sentence embedding of size 32x24. The two level system has 178 126 parameters.

\newpage

\subsection{Sentence-level analysis}
\label{sec:confmats}

\subsubsection{Sample confusion matrices for altered latent values}

\begin{figure}[h]
    \centering
    \includegraphics[width=\textwidth]{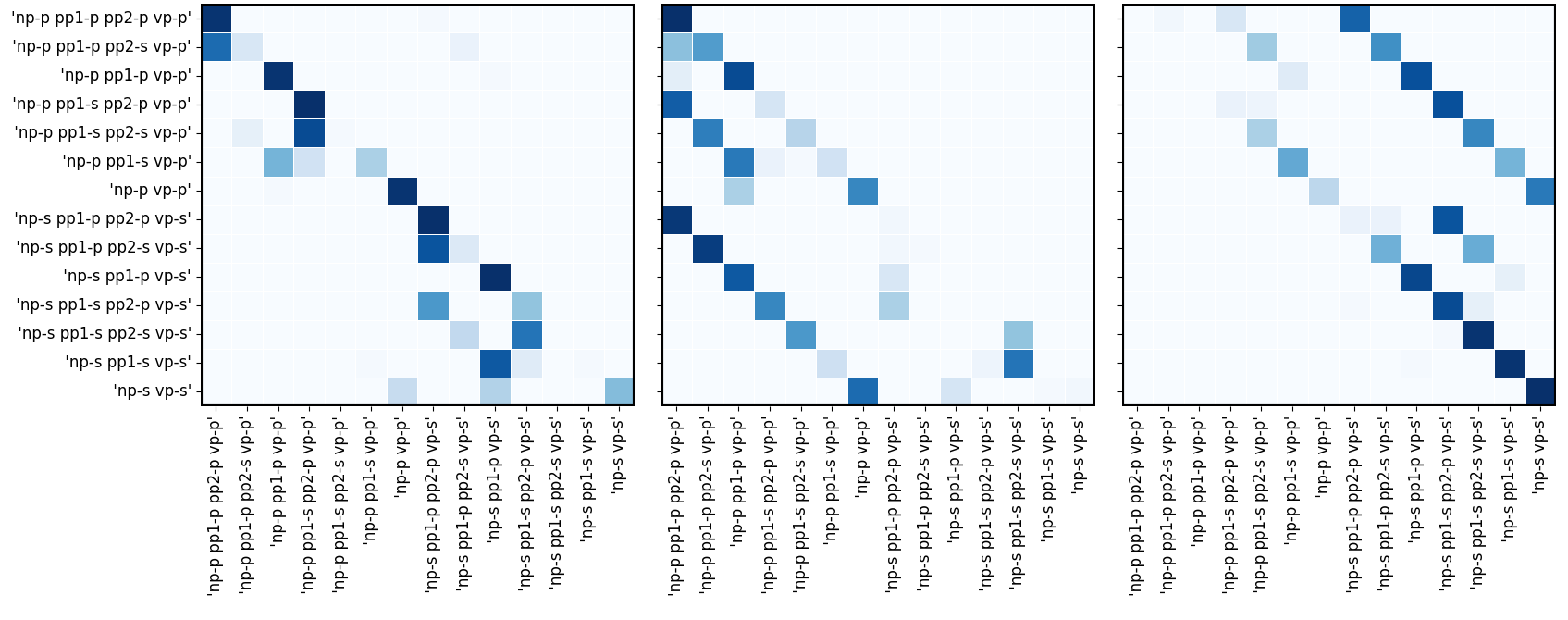}
    \caption{Confusion matrices for altered values on units 1 (left matrix), unit 2 (middle matrix) and unit 3 (right matrix)}
    \label{fig:confmats}
\end{figure}

Each matrix shows a particular way of conflating different patterns:
\begin{itemize}
    \item changes to values in unit 1 of the latent lead to patterns that differ in the grammatical number of $pp2$ to become indistinguishable
    \item changes to values in units 2 and 3 of the latent lead to the conflation of patterns that have different subject-verb numbers.
\end{itemize}

\subsubsection{Sentence-level analysis for English data}

\begin{minipage}{\textwidth}
    \begin{minipage}{0.5\textwidth}
        \centering
        \includegraphics[scale=0.25,trim={0 0 0 2.1cm},clip]{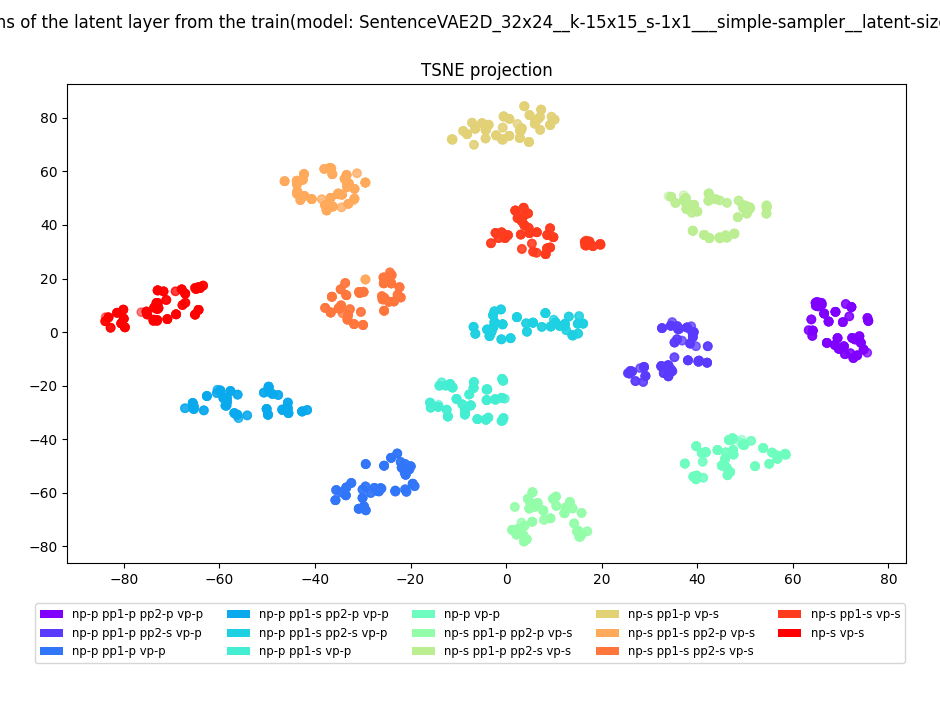}            
    \end{minipage}
    \hfill
    \begin{minipage}{0.5\textwidth}
        \centering
        \includegraphics[scale=0.2]{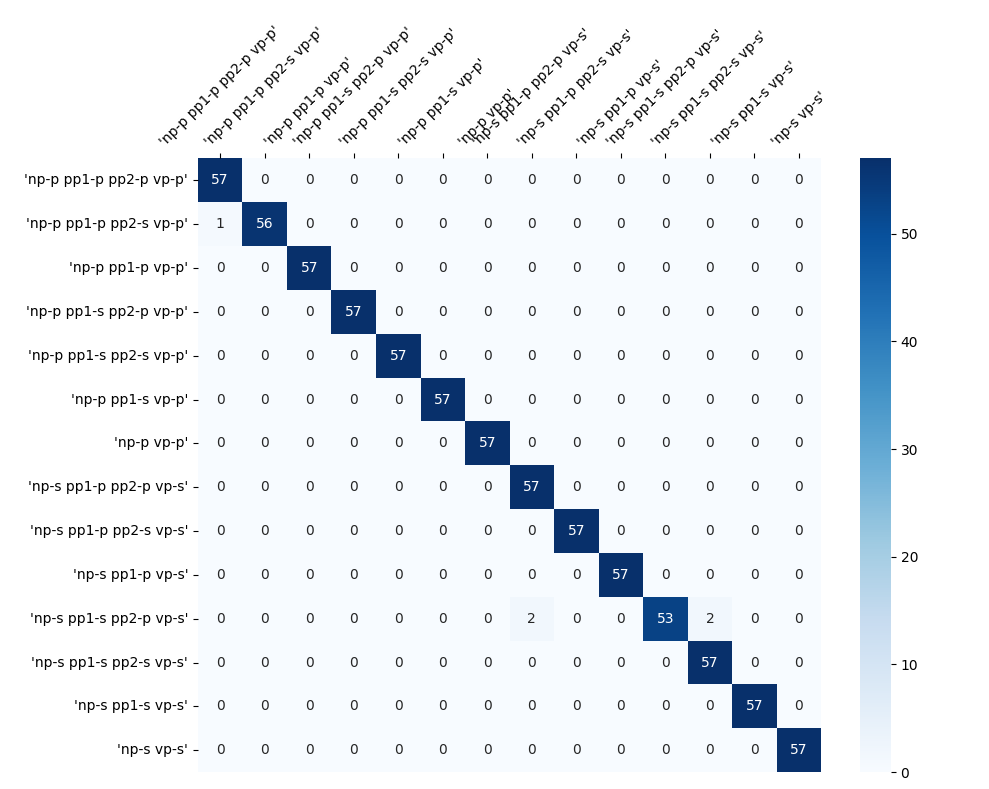}
    \end{minipage}
    \captionof{figure}{Chunk identification results: tSNE projections of the latent vectors for the English dataset, and confusion matrix of the system output.}
    \label{fig:partsinsentences_EN}
\end{minipage}

\newpage

\subsection{Detailed task results}

\begin{table}[h]
    \centering
    \begin{tabular}{lllc} \\ 
{\sc train on} & {\sc test on} & {\sc VAE} & {\sc 2 level VAE} \\ \hline \\
 \multicolumn{4}{l}{\bf BLM agreement} \\ \hline
\ti   &   \ti & 0.929 (0)  & {\bf 0.935} (0.0049) \\ \cline{3-4} 
\ti   &  \tii & 0.899 (0)  & {\bf 0.908}	(0.0059) \\ \cline{3-4}
\ti   & \tiii & 0.662 (0)  & {\bf 0.871}	(0.0092) \\ \cline{2-4}
\tii  &   \ti & 0.948 ($<$e-10)  & {\bf 0.974} (0.0049) \\ \cline{3-4}
\tii  &  \tii & 0.879 ($<$e-10)  & {\bf 0.904} (0.0021) \\ \cline{3-4}
\tii  & \tiii & 0.713 (0)  & {\bf 0.891}	(0.0015) \\ \cline{2-4}
\tiii &   \ti & {\bf 0.851} (0.037)  & 0.611 (0.1268) \\ \cline{3-4}
\tiii &  \tii & {\bf 0.815} (0.0308)  & 0.620 (0.1304) \\ \cline{3-4}
\tiii & \tiii & {\bf 0.779} (0.0285)  & 0.602 (0.1195) \\ \hline \\
    \multicolumn{4}{l}{\bf BLM verb alternation group 1} \\ \hline
\ti   &   \ti & 0.989 (0)  & {\bf 0.995} ($<$e-10) \\ \cline{3-4} 
\ti   &  \tii & 0.907 (0)  & {\bf 0.912}	(0.0141) \\ \cline{3-4}
\ti   & \tiii & {\bf 0.809} (0)  & 0.804	(0.0167) \\ \cline{2-4}
\tii  &   \ti & 0.989 (0)  & {\bf 0.996} (0.0013) \\ \cline{3-4}
\tii  &  \tii & 0.979 ($<$e-10)  & {\bf 0.984} (0.0016) \\ \cline{3-4}
\tii  & \tiii & 0.915 (0)  & {\bf 0.928}	(0.0178) \\ \cline{2-4}
\tiii &   \ti & 0.997 (0)  & {\bf 0.999} (0.0013) \\ \cline{3-4}
\tiii &  \tii & 0.977 (0)  & {\bf 0.986} (0.0027) \\ \cline{3-4}
\tiii & \tiii & 0.98 (0)  & {\bf 0.989} (0.0003) \\ \hline
\\
    \multicolumn{4}{l}{\bf BLM verb alternation group 2} \\ \hline
\ti   &   \ti & {\bf 0.992} (0)  & 0.987 (0.0033) \\ \cline{3-4} 
\ti   &  \tii & 0.911 (0)  & {\bf 0.931}	(0.0065) \\ \cline{3-4}
\ti   & \tiii & 0.847 (0)  & {\bf 0.869}	(0.0102) \\ \cline{2-4}
\tii  &   \ti & {\bf 0.997} (0)  & 0.993 (0.0025) \\ \cline{3-4}
\tii  &  \tii & {\bf 0.978} ($<$e-10)  & {\bf 0.978} (0.0017) \\ \cline{3-4}
\tii  & \tiii & 0.923 (0)  & {\bf 0.956}	(0.0023) \\ \cline{2-4}
\tiii &   \ti & 0.979 ($<$e-10)  & {\bf 0.981} (0.0022) \\ \cline{3-4}
\tiii &  \tii & 0.972 (0)  & {\bf 0.975} (0.0005) \\ \cline{3-4}
\tiii & \tiii & 0.967 (0)  & {\bf 0.977} (0.0022) \\ \hline
\end{tabular}
\caption{Analysis of systems: average F1 (std) scores (over 3 runs) for the VAE and 2xVAE systems. The highest value for each train/test combination highlighted in bold.}
\label{tab:sys_analysis_1d}
\end{table}

\newpage

\subsection{Detailed error results}

    \begin{figure}[h]
        \centering
        \includegraphics[scale=0.3]{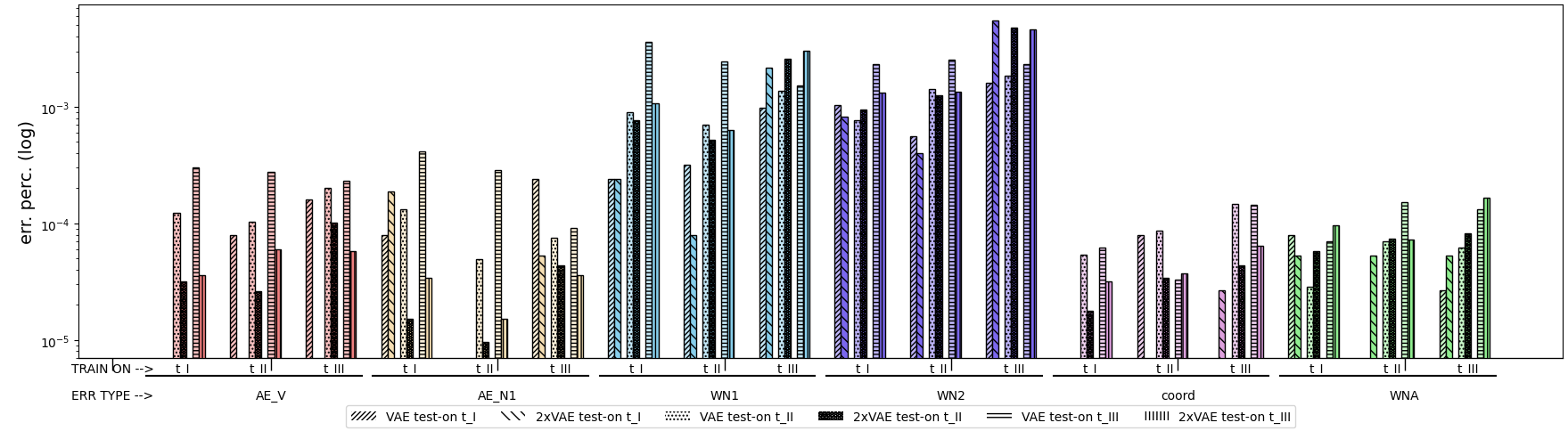}
        \caption{Agreement error analysis: y-axis is the log of error percentages. N1\_alter and N2\_alter are sequence errors.}
        \label{fig:agreementErrs}
        \end{figure}

    \begin{figure}[h]
        \centering
        \includegraphics[scale=0.3]{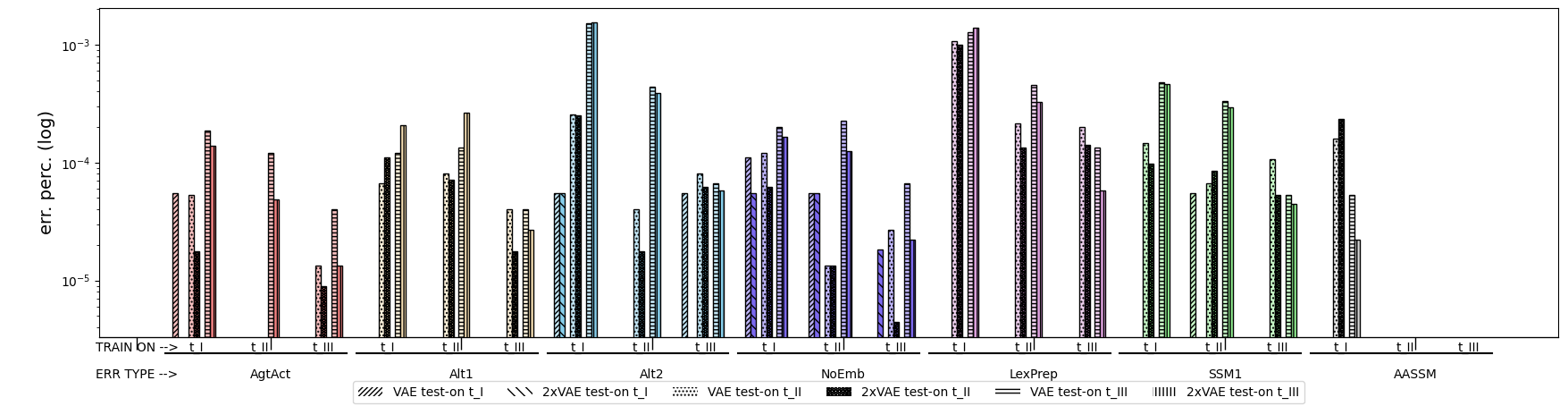}
        \caption{Verb alternation group1 error analysis: y-axis is the log of error percentages. }
        \label{fig:agreementErrs-s/lE}
    \end{figure}

    \begin{figure}[h]
        \centering
        \includegraphics[scale=0.3]{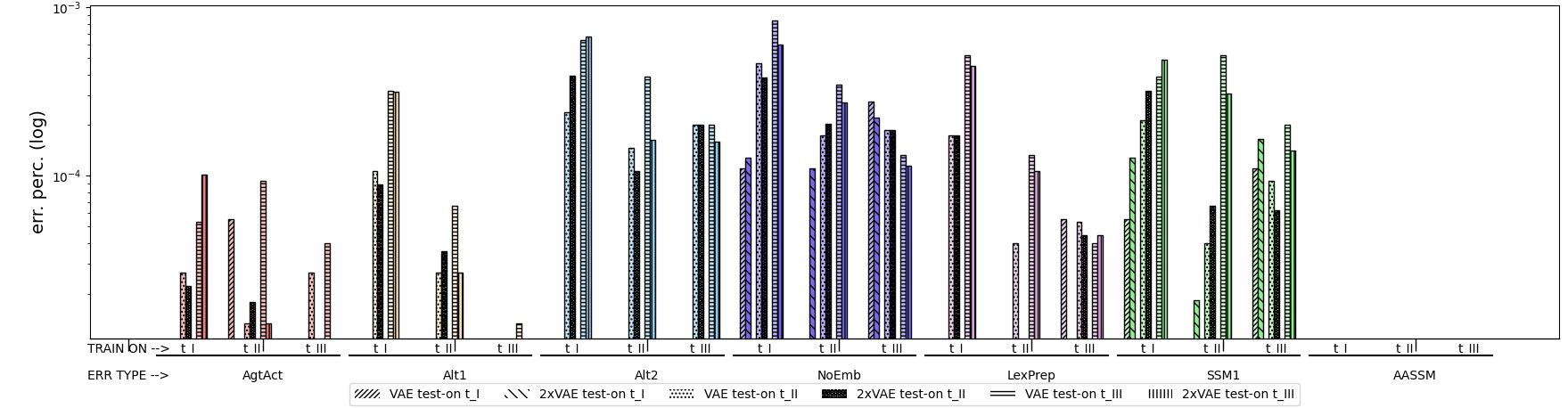}
        \caption{Verb alternation group2 error analysis: y-axis is the log of error percentages. }
        \label{fig:agreementErrs-s/lE}
    \end{figure}

\end{document}